\documentclass[nolinenumbers]{iucrjournals}
\usepackage{amsmath,amssymb,amsfonts}
\usepackage{algorithmic}
\usepackage{algorithm}
\usepackage{array}
\usepackage{bm}
\usepackage{textcomp}
\usepackage{stfloats}
\usepackage{url}
\usepackage{verbatim}
\usepackage{graphicx}
\usepackage{textcomp}
\usepackage{xcolor}
\usepackage{dsfont}
\usepackage{bm,mathabx,amsthm}
\usepackage{subcaption}
\usepackage{graphicx}
\usepackage{booktabs}
\usepackage{romannum}

\usepackage{bm}
\usepackage{microtype}
\usepackage{graphicx}
\usepackage{subcaption}
\usepackage{booktabs} 

\usepackage[capitalize,noabbrev]{cleveref}
\usepackage[textsize=tiny]{todonotes}

\usepackage{threeparttable}
\usepackage{wrapfig}
\usepackage{multicol}
\usepackage{multirow}
\usepackage{makecell}

\usepackage{titletoc}
\usepackage[page,header]{appendix}
\newcommand{\revise}[1]{{#1}}
\title{From Solo to Symphony: Orchestrating Multi-Agent Collaboration with Single-Agent Demos}

\author{Xun Wang}
\author{Zhuoran Li}
\author{Yanshan Lin}
\author{Hai Zhong}
\author{Longbo Huang\thanks{Corresponding Author.}}

\affil{Institute for Interdisciplinary Information Sciences (IIIS), Tsinghua University \{wang-x24,lizr20,linys23,zhongh22\}@mails.tsinghua.edu.cn, longbohuang@tsinghua.edu.cn }

\begin{document} 
\maketitle 
\makeatletter

\begin{abstract}
  Training a team of agents from scratch in multi-agent reinforcement learning (MARL) is highly inefficient, much like asking beginners to play a symphony together without first practicing solo. Existing methods, such as offline or transferable MARL, can ease this burden, but they still rely on costly multi-agent data, which often becomes the bottleneck. In contrast, solo experiences are far easier to obtain in many important scenarios, e.g., collaborative coding, household cooperation, and search-and-rescue. To unlock their potential, we propose Solo-to-Collaborative RL (SoCo), a framework that transfers solo knowledge into cooperative learning. SoCo first pretrains a shared solo policy from solo demonstrations, then adapts it for cooperation during multi-agent training through a policy fusion mechanism that combines an MoE-like gating selector and an action editor. Experiments across diverse cooperative tasks show that SoCo significantly boosts the training efficiency and performance of backbone algorithms. These results demonstrate that solo demonstrations provide a scalable and effective complement to multi-agent data, making cooperative learning more practical and broadly applicable.
\end{abstract}


\section{Introduction}
Multi-agent reinforcement learning (MARL) has emerged as a core paradigm for sequential decision making in environments that require coordination \cite{shoham2008multiagent, lowe2017multi, gronauer2022multi}. By interacting with the environment and receiving feedback, MARL enables agents to learn cooperative policies, providing a principled framework for solving complex decision-making problems such as autonomous driving \cite{zhang2024multi}, large-scale network optimization \cite{stepanov2024fair}, and collaborative robotics \cite{tang2025deep}. 

However, compared to single-agent RL, MARL faces inherent challenges \cite{busoniu2008comprehensive,hernandez2019survey}, including dimensionality explosion, coordination difficulty, and environmental non-stationarity. As a result, training joint policies from scratch is often inefficient, much like asking novices to rehearse a symphony without prior practice: difficult, time-consuming, and unlikely to yield good results. This inefficiency poses a major obstacle to applying MARL effectively in practice.

To address these challenges, a growing line of research has explored offline MARL \cite{omar, shao2023counterfactual, li2023beyond,liu2024offlinemultiagentreinforcementlearning} and offline-to-online fine-tuning \cite{zhong2025offline}. These methods learn from pre-collected task-specific cooperative trajectories to avoid costly exploration, and refine pretrained policies with limited online rollouts when interaction is allowed. More recent studies have attempted to relax the data assumption by leveraging multi-task cooperative datasets \cite{zhang2023odis, chen2024variational, hissd} or even non-cooperative multi-agent datasets \cite{wang2023dm2, yu2025beyond}. These efforts broaden the scope of usable data and represent important progress, but they remain fundamentally tied to multi-agent trajectories.

Actually, in many cooperative problems, there often exists a corresponding solo version whose demonstrations are much easier to obtain and learn from. \revise{For example, in collaborative coding \cite{coding} a single coder writes a short piece of code, in household cooperation \cite{household} a single robot performs an individual chore, and in search-and-rescue \cite{sar} a single drone searches for a target.} Although such demonstrations deviate from the target cooperative setting, they are far from useless. Such as in orchestral performance, it is more effective to let each novice player first master the basics of solo play before attempting a full ensemble. Yet, the potential of solo demonstrations to accelerate MARL remains underexplored. This gap motivates an important but underexplored question:
\begin{center}
    \textbf{\textit{Can solo demonstrations be effectively leveraged to accelerate the collaborative MARL?}}
\end{center}
An affirmative answer to this hypothesis will validate solo data as a scalable and cost-effective resource. This will be instrumental in fostering efficient learning in settings where cooperative data are limited but solo demonstrations are plentiful \revise{\cite{household,sar,coding}}, consequently making MARL a more viable solution for practical applications. 

However, addressing this problem is non-trivial and involves two major challenges. \revise{The first is \textit{observation mismatch}: differences in observation dimensionality hinder the direct transfer of solo demonstrations to multi-agent training \cite{hu2021updet, zhang2023odis, hissd}. In some cases, a single local observation may even correspond to multiple distinct solo views, creating ambiguity for policy reuse. The second is \textit{domain shift}: unlike multi-agent data, whether joint or agent-specific, that inherently encode cooperation \cite{wang2023dm2,yu2025beyond}, solo data contain no such information. In addition, discrepancies in environment dynamics between solo and cooperative settings (e.g., individual attributes and observation noise) further exacerbate the gap. These challenges hinder direct policy transfer, highlighting the need to distill knowledge from solo demonstrations and integrate it into cooperative learning.}

To tackle these challenges, we propose \textbf{\underline{So}lo-to-\underline{Co}llaborative RL} (SoCo) framework, which transfers knowledge from solo demonstrations to cooperative MARL. SoCo first pretrains a shared solo policy from solo demonstrations via imitation learning, providing a common skill prior for all agents. During cooperative training, local observations are decomposed into solo views aligned with the demonstrations, allowing the reuse of the solo policy to obtain candidate actions. Then, a policy fusion module \revise{selects and refines these actions for each agent, adapting them to the cooperative setting and mitigating domain shift.} Specifically, \revise{inspired by MoE \cite{moe} and action fusion in single-agent RL \cite{dong2025expo}}, a learnable gating selector chooses the most suitable candidate, while an action editor refines it for effective cooperation. This design not only tackles the challenge of solo-to-cooperative transfer but also provides flexibility for task-specific customization.

We validate SoCo across diverse cooperative benchmarks, and the results show that it markedly improves training efficiency while achieving competitive or superior performance. These findings highlight the potential of solo demonstrations as a scalable resource for cooperative MARL.

Our main contributions are summarized as follows:
\begin{itemize}
\item We investigate an important yet underexplored problem of leveraging solo demonstrations to benefit cooperative MARL, and show that such data, though lacking explicit cooperative information, can substantially accelerate multi-agent training.
\item \revise{We develop Solo-to-Collaborative RL (SoCo), a framework for solo-to-cooperative transfer. It decomposes local observations to reuse a pretrained shared solo policy, and employs a policy fusion module trained from cooperative interactions that combines a {gating selector} for choosing solo actions with an {action editor} for refining them, enabling more efficient cooperation.}
\item \revise{We validate SoCo on cooperative benchmarks with diverse characteristics and difficulty, showing that it effectively addresses observation ambiguity and domain shift, boosts training efficiency, and achieves competitive or superior performance}, highlighting the potential of solo demonstrations as a scalable resource for MARL. 
\end{itemize}

\section{Preliminary}
\subsection{Multi-Agent Reinforcement Learning}
We model multi-agent reinforcement learning (MARL) within the decentralized partially observable Markov decision process (Dec-POMDP) framework \cite{oliehoek2016concise}. Formally, a Dec-POMDP is defined as $\mathcal{M} = \langle \mathcal{N}, \mathcal{S}, \mathcal{A}, \mathcal{O}, P, R, \gamma \rangle,$
where $\mathcal{N} = \{1, \dots, N\}$ denotes the set of agents, $\mathcal{S}$ is the global state space, and $\mathcal{A}$ and $\mathcal{O}$ are the joint action and observation spaces, each formed from the agents’ local action $\{\mathcal{A}_i\}_{i=1}^{N}$ and observation spaces $\{\mathcal{O}_i\}_{i=1}^{N}$. At each time step, every agent receives a local observation generated from the current global state. Based on the joint action $\mathbf{a} = (a_1,\dots,a_N)$, the environment transitions to the next state according to $P$, and a shared reward $R(s,\mathbf{a})$ is returned. The goal of MARL is to learn a joint policy $\boldsymbol{\pi} = (\pi_1, \dots, \pi_N)$ that maximizes the expected discounted return:$J(\boldsymbol{\pi}) = \mathbb{E}_{\boldsymbol{\pi}} \!\left[ \sum_{t=0}^{\infty} \gamma^t R(s_t, \boldsymbol{a}_t) \right]$. The solo case naturally arises when $|\mathcal{N}|=1$. 

\subsubsection{CTDE Paradigm and Deterministic Policy Gradient Method}
Centralized Training with Decentralized Execution (CTDE) \cite{oliehoek2008optimal,IntroCTDE,ctde_new} is a widely adopted paradigm in cooperative MARL. In CTDE, each agent executes its policy in a decentralized manner, relying only on its own local observation during interaction with the environment. During training, however, the learning process can leverage additional global information (e.g., global states or joint actions) through centralized critics. This design improves training stability and coordination, while keeping execution scalable and realistic.

A representative CTDE algorithm is MADDPG \cite{lowe2017multi}. It extends the deterministic policy gradient (DPG) framework to multi-agent settings by introducing a centralized critic for each agent, while keeping actors decentralized. Formally, let agent $i$ have policy $\pi_i(o_i;\theta_i)$ parameterized by $\theta_i$, and the deterministic policy gradient for agent $i$ is:
$$
\nabla_{\theta_i} J(\pi_i) \approx 
\mathbb{E}_{s,\mathbf{a} \sim \mathcal{D}} 
\big[ \nabla_{\theta_i} \pi_i(o_i) \, 
\nabla_{a_i} Q_i(s, \boldsymbol{a}; \psi^i) \big|_{a_i=\pi_i(o_i)} \big],
$$
{where $Q_i(s,\boldsymbol{a}; \psi^i)$ denotes the centralized critic for agent $i$, parameterized by $\psi^i$. In practice, however, the critics often share a single parameter set $\psi$} across agents, whereas each agent maintains its own policy network.

Building on MADDPG, MATD3 \cite{matd3} incorporates the improvements of TD3 \cite{td3}, including twin critics, target smoothing, and delayed policy updates. HATD3 \cite{harl} further extends MATD3 by introducing a heterogeneous sequential optimization. In this paper, we focus primarily on the DPG family under the CTDE paradigm, as represented by MATD3 and HATD3. Nevertheless, the proposed framework is, in principle, extendable to stochastic policy methods, {such as MAPPO \cite{yu2021surprising} or HASAC \cite{liu2024maximum}.}

\section{Solo-to-Collaborative RL}
To bridge the gap between solo demonstrations and multi-agent cooperation, and to tackle observation mismatch and domain shift, we propose the Solo-to-Collaborative RL (SoCo) framework. SoCo first learns a shared solo policy from solo demonstrations. Then, during cooperative training, local observations are decomposed into solo views, allowing the reuse of the solo policy to obtain candidate actions. Finally, {a per-agent policy-fusion module selects the most appropriate candidate policy and refines it for each agent, adapting it to the cooperative setting and mitigating domain shift. }
In the following, we present each component in detail. The full algorithm is presented in Algorithm \ref{alg:s2m} in Appendix \ref{app:algo}.

\subsection{Solo Policy Extraction}
A solo policy is learned by imitation from the demonstration dataset $\mathcal{D}_s$ and shared by all the agents. For simplicity, our implementation uses standard behavior cloning, minimizing the mean-squared error between the policy’s action and the action recorded in $\mathcal{D}_s$ to obtain a deterministic behavioral policy:
\begin{equation}\label{eq:bc_loss}
    \min_{w}\ \mathbb{E}_{(o,a)\sim \mathcal{D}_s}\bigl\|\beta_{w}(o)-a\bigr\|_2^2.
\end{equation}

This design choice is flexible rather than mandatory: one could instead adopt a stochastic imitation model that learns $\beta_w(a \mid o)$, for example, parameterizations based on VAE \cite{kingma2013auto}, diffusion models \cite{yang2023diffusion}, or flow matching \cite{lipman2022flow}, by simply switching to a likelihood-based objective without altering the subsequent components of the framework. Finally, the solo policy is shared across agents, and its parameters are frozen during the cooperative learning phase.

\begin{figure}[!tpb]
\centering
\centerline{\includegraphics[width=1.0\textwidth]{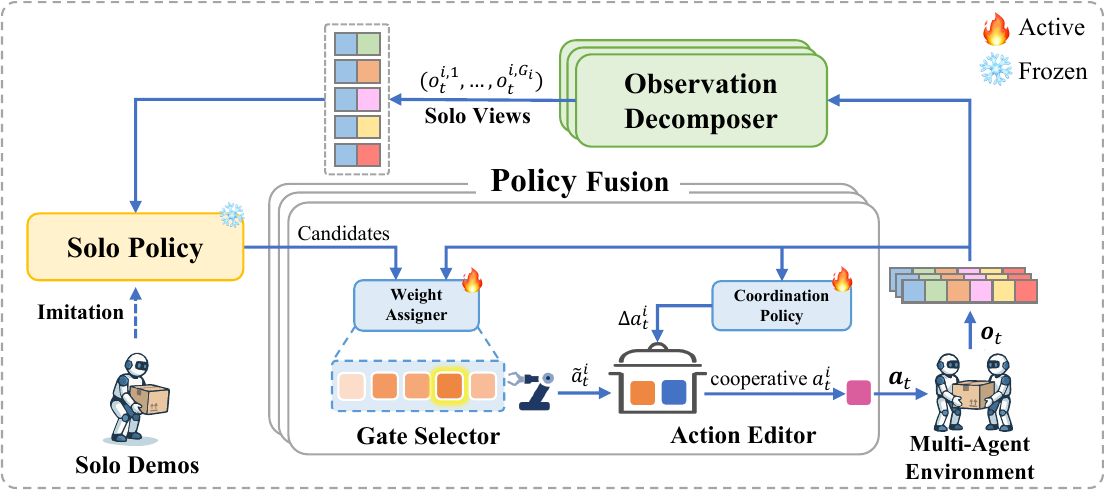}}
\caption{SoCo framework. \revise{A shared solo policy is pretrained from demonstrations and kept frozen, then reused through observation decomposition during cooperative training. Coordination ability is injected by the Policy Fusion module, where the Gating Selector selects suitable solo actions and the Action Editor fine-tunes them to mitigate domain shift.}
} 
\label{fig:s2m}
\end{figure}
\subsection{Observation Decomposition}\label{sec:decompose}

Given the settings of the cooperative tasks and their corresponding solo tasks considered in this paper, we make a reasonable assumption that the observations in these cooperative tasks are well-defined, structured, and decomposable. Specifically, each observation consists of own features (e.g., velocity, position) and stacked features of all other entities (e.g., teammates or target states). \revise{The observation space of the corresponding solo tasks can then be constructed from these feature units (e.g., controlling one HalfCheetah vs. multiple coupled HalfCheetahs).}

Hence, following prior works \cite{pmlr-v100-liu20a, wu2024portal, hissd}, we introduce a rule-based observation decomposer. Concretely, we decompose the observation of the $i$-th agent at time step $t$, denoted as $o^i_t$, into the self-related component $o_t^{i,0}$ and the entity-related components $\{o_t^{i,k}\}_{k=1}^{K_i}$, where $K_i$ denotes the total number of entities observable by agent $i$. When deployed in cooperative environments, depending on the specific task, we may reassemble the decomposed feature units into $G_i$ valid solo views $\{\tilde{o}_t^{i,k}\}_{k=0}^{G_i}$ for agent $i$ by concatenation, zero-padding, and so on, thereby addressing the issue of inconsistent observation spaces. 

\subsection{Policy Fusion}
In the cooperative training phase, the pretrained solo policy cannot be directly transferred. The main obstacles are twofold: (i) \revise{a single local observation may map to multiple solo views, producing several candidate actions (e.g., toward different targets) that must be disambiguated}; and (ii) domain shift between solo and multi-agent settings often degrades performance, necessitating fine-tuning for effective adaptation.

Therefore, inspired by Mixture-of-Experts (MoE) \cite{moe} and action fusion techniques in single-agent RL \cite{dong2025expo}, we propose a novel learnable policy fusion module. Notably, our design operates at the agent level and is trained directly on multi-agent samples with standard MARL optimization, thereby injecting cooperative adaptability into solo policies. Within this module, each agent employs a \textit{Gating Selector} to resolve ambiguity by selecting suitable solo actions, and an \textit{Action Editor} to fine-tune the chosen action for coordination, together enabling effective solo-to-cooperative transfer.
\subsubsection{Gating Selector} 
As discussed in Section \ref{sec:decompose}, the local observation $o_t^i$ of the $i$-th agent corresponds to $G_i$ solo views $\{\tilde{o}_t^{i,k}\}_{k=0}^{G_i}$. By applying the solo policy, these yield $G_i$ candidate actions:
\begin{equation}
a_t^{i,k} = \beta(\tilde{o}_t^{i,k}), \quad k=1,\dots,G_i .
\end{equation}

However, due to the solo-to-cooperative gap, not all candidate actions are suitable for the current cooperative context, and some may even conflict with each other. To resolve this, SoCo equips each agent with a \revise{weight assigner} $g^i_\varphi: \mathcal{O}_i \rightarrow \mathbb{R}^{G_i}$ that, conditioned on the current local observation $o_t^i$, evaluates the candidate solo actions and assigns weights to them, thereby selecting the most appropriate one for coordination.

To enable learnable action selection, we adopt the Gumbel–Softmax reparameterization \cite{jang2017categorical} with the straight-through estimator. The gating weights $g^i_\varphi(o_t^i)$ define a categorical distribution, from which a one-hot action is drawn: the forward pass takes the most probable action, while the backward pass propagates gradients through the soft sample. The resulting action for agent $i$ is:
\begin{equation}\label{eq:gate}
\tilde{a}_t^i \;=\; \left\langle \mathrm{GumbelSoftmax}\bigl(g^i_\varphi(o_t^i)\bigr),~ \boldsymbol{a}_t^i \right\rangle ,
\end{equation}

where $\boldsymbol{a}_t^i = (a_t^{i,1}, \dots, a_t^{i,G_i})$ is the set of candidate actions derived from the solo policy.

Moreover, this module is designed to be both general and flexible, allowing adaptation to different scenarios. For instance, the gating function may be rule-based instead of learned, and in the special case of $G_i=1$, the selector can be omitted entirely.

\subsubsection{Action Editor} \label{sec:ae}

To leverage the prior knowledge in solo actions while overcoming transfer difficulties from domain shift, we design an action editor that injects cooperative information through residual corrections. \revise{Specifically, we introduce a coordination policy $\pi_\theta:\mathcal{O}_i \rightarrow \mathcal{A}_i$ that produces a raw residual adjustment to the solo action. To keep this correction bounded and scale-invariant while avoiding gradient saturation, we squash the policy output with $f_L(x)=L\,\tanh(x/L)$. Given the current local observation $o_t^i$, the adjustment is:} 
\begin{equation}\label{eq:aedit}
\Delta a_t^i \;=\; \begin{cases}
    L \cdot \tanh\!\left(\dfrac{\pi_\theta(o_t^i)}{L}\right)& \text{if } L>0\\
    \boldsymbol{0} & \text{if }L=0
    \end{cases}
\end{equation}
where $L$ is a hyperparameter that controls the strength of the correction. By tuning $L$, the framework can trade off between leveraging solo priors and adapting to multi-agent dynamics. 

Then, the final cooperative action is defined as: 
\begin{equation}\label{eq:comb}
\revise{a_t^i \;=\;\text{Clip}\bigl(\tilde{a}_t^i + \Delta a_t^i \bigr)}
\end{equation}
where $\tilde{a}_t^i$ denotes the solo action selected by the gating selector, and $\text{Clip}(\cdot)$ is a clipping operator to prevent action overflow. In our implementation, we adopt a $\tanh$-based operator, but the design is modular and allows substituting other operators depending on the task.

\subsection{Collaborative Policy Optimization}

For notational simplicity, we denote the fused policy as $\Pi_\phi$, where $\phi = \{\varphi, \theta\}$ collects the learnable parameters of the gating selector and action editor. Since SoCo is fully decoupled from the backbone algorithm, this policy can be optimized with any MARL method; here we instantiate it with MATD3 \cite{matd3} for concreteness.

Each agent $i$ maintains two shared centralized critics $Q_{\psi_1}, Q_{\psi_2}$ and an individual actor $\Pi_{\phi_i}$. Given a batch of transition $\mathcal{B}=\{(s_t, \boldsymbol{o}_t,\boldsymbol{a}_t,r_t,s_{t+1},\boldsymbol{o}_{t+1})\}$, the critic loss is:
\begin{equation}\label{eq:qloss}
\mathcal{L}(\psi_j) = 
\mathbb{E}_{(s_t,\boldsymbol{o}_t,\boldsymbol{a}_t,r_t,\boldsymbol{o}_{t+1}) \sim \mathcal{B}}
\Bigl[\bigl(Q_{\psi_j}(s_t, \boldsymbol{a}_t) - y_t\bigr)^2\Bigr], \quad j=1,2,
\end{equation}
where the target $y_t$ is defined as
\begin{equation}
y_t = r_t + \gamma \, 
\min_{k=1,2} Q_{\bar{\psi}_k}\bigl(s_{t+1}, \boldsymbol{a}_{t+1}'\bigr), 
\quad 
(\boldsymbol{a}_{t+1}')_i = \Pi_{\bar{\phi}_i}(o_{t+1}^i) + \epsilon, 
\end{equation}
with $\{\bar{\psi}_k\}_{k=1}^2$ and $\bar{\phi}_i$ denoting target networks and $\epsilon$ being clipped Gaussian noise for policy smoothing. The actors are optimized by maximizing the Q-value estimated by the first critic:
\begin{equation}\label{eq:ploss}
\mathcal{L}(\phi_i) = 
- \mathbb{E}_{(s_t,\boldsymbol{o}_t) \sim \mathcal{B}}
\Bigl[ Q_{\psi_1}\bigl(s_t, \Pi_{\phi_i}(o_t^i)\bigr) \Bigr].
\end{equation}
For algorithms with stochastic policies, it suffices to compute the log-probability of the fused action according to Eq. (\ref{eq:comb}) and substitute it into the policy loss.

{In this way, SoCo provides a plug-and-play bridge between solo demonstrations and cooperative learning, turning single-agent demonstrations into a scalable and effective complement to multi-agent data, making cooperative learning more practical and broadly applicable.}

\section{Experiments}\label{sec:exp}
To evaluate the proposed SoCo framework, we conduct experiments on a variety of cooperative tasks. Our goals are to investigate the following questions: (i) Can SoCo improve the sample efficiency of multi-agent algorithms? (ii) Can SoCo enhance the ultimate performance of multi-agent algorithms? (iii) How do the individual components of SoCo contribute to its effectiveness?
\subsection{Setup}
\paragraph{Environments and Tasks.}Following prior works \cite{use_walker,use_mpe, use_mamujoco}, our experiments cover nine tasks from four representative cooperative scenarios: (i) \textit{Spread} \cite{lowe2017multi,terry2021pettingzoo}, where agents must cover distinct landmarks. This setting introduces target ambiguity but involves little domain shift. (ii) \textit{LongSwimmer} \cite{facmac,gymnasium_robotics2023github}, where a multi-segment worm must swim forward, with each agent controlling two consecutive joints. These tasks do not involve target ambiguity but introduce domain shift due to altered dynamics. (iii) \textit{MultiHalfCheetah} \cite{facmac,gymnasium_robotics2023github}, where multiple HalfCheetahs are connected in a chain and must run forward together. These tasks avoid target ambiguity but involve noticeable domain shift and present a non-trivial control challenge. (iv) \textit{MultiWalker} \cite{gupta2017cooperative,terry2021pettingzoo}, where multiple bipedal robots jointly carry a package forward. These tasks avoid target ambiguity, but are inherently very difficult, with severe domain shift on top of coordination challenges. 

Considering the characteristics and difficulty of these environments, we evaluate tasks with 3, 4, and 5 agents in \textit{Spread} and \textit{LongSwimmer}, 2 and 3 agents in \textit{MultiHalfCheetah}, and 2 agents in \textit{MultiWalker}. Details on these environments and tasks are provided in Appendix \ref{app:env}.

\paragraph{Data Collection.}To collect solo demonstration data, we first train policies with TD3 \cite{td3} on the corresponding solo tasks until convergence, and then record $1$M transition samples. The corresponding solo tasks are: (i) a single agent navigating to one landmark, (ii) a 3-segment worm with 2 joints swimming forward, (iii) a HalfCheetah with altered attributes running forward, and (iv) a single bipedal robot carrying the long package forward. When transferred to multi-agent settings, these lead to (i) goal ambiguity, (ii) domain shift, (iii) notable domain shift and cooperative difficulty, and (iv) severe domain shift coupled with substantial cooperative difficulty. More details can be found in Appendix \ref{app:solo}.

\paragraph{Baselines.}We adopt two representative DPG algorithms, MATD3 \cite{matd3} and HATD3 \cite{harl}, as our baselines. Their implementations are taken from the HARL codebase \cite{harl}, on top of which we also build our SoCo variants.

\paragraph{Experiment Setup.} SoCo first undergoes imitation learning to obtain the solo policy ($5$k steps for \textit{Spread}, $100$k steps for the others). During cooperative learning, all algorithms perform $10$k random interaction steps for warm-up, followed by policy optimization with the same number of environment steps ($2$M steps for \textit{LongSwimmer} and \textit{MultiHalfCheetah}, $5$M steps for the others). Except for the correction strength $L$ in SoCo, all hyperparameters are identical to the default settings. For evaluation, each trained policy is tested over $40$ episodes, and the average return is reported. All experiments are repeated with $3$ random seeds to account for variance. Detailed hyperparameter settings are provided in Appendix \ref{app:hyper}.
\subsection{Evaluation Results and Analysis}
\begin{figure}[!t]
\centering
    \begin{subfigure}[b]{0.3\columnwidth}
        \centering
        \includegraphics[width=.9\textwidth]{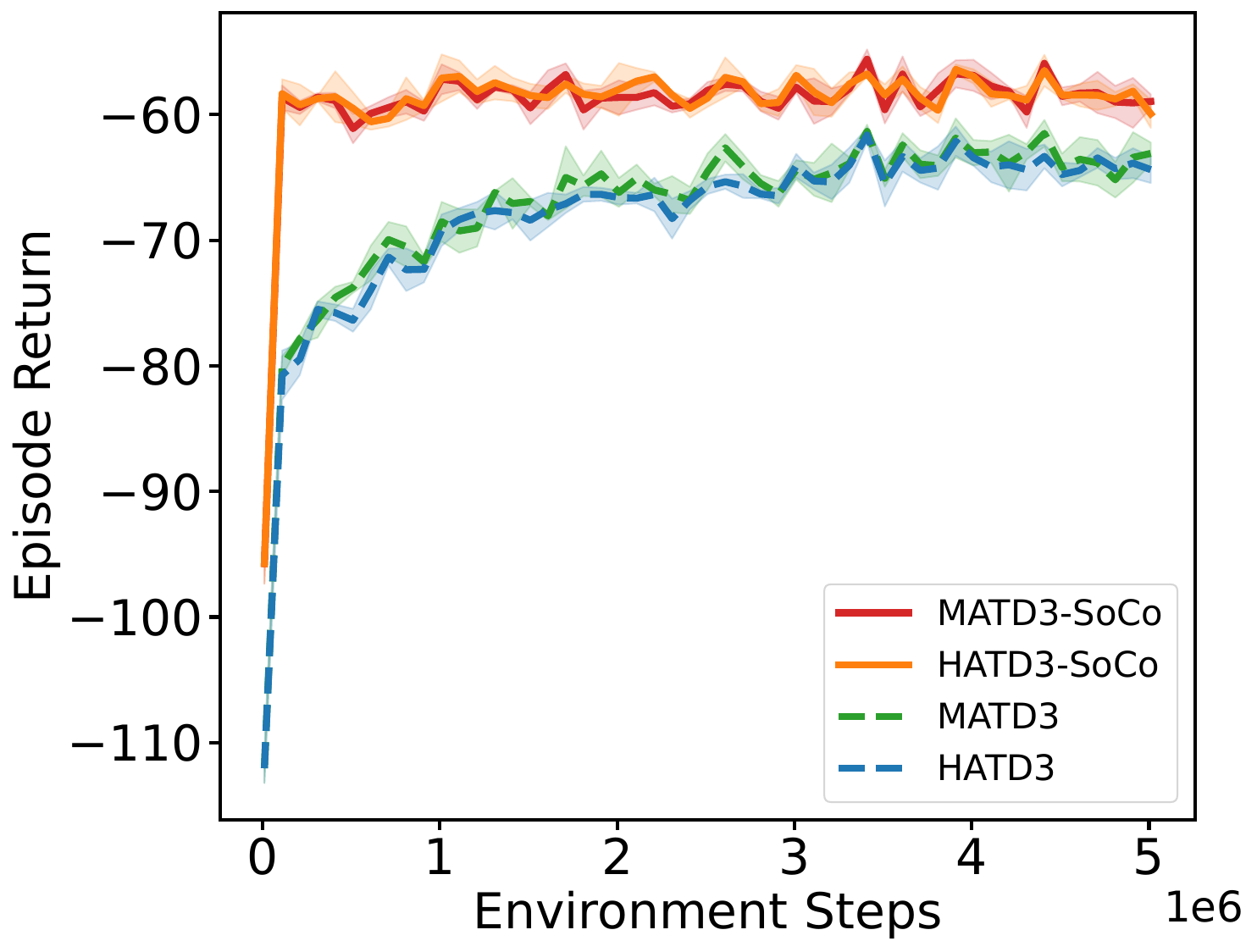}
        \caption{Spread (3 agents).}\label{fig:s3}
    \end{subfigure}
    \begin{subfigure}[b]{0.3\columnwidth}
        \centering
        \includegraphics[width=.9\textwidth]{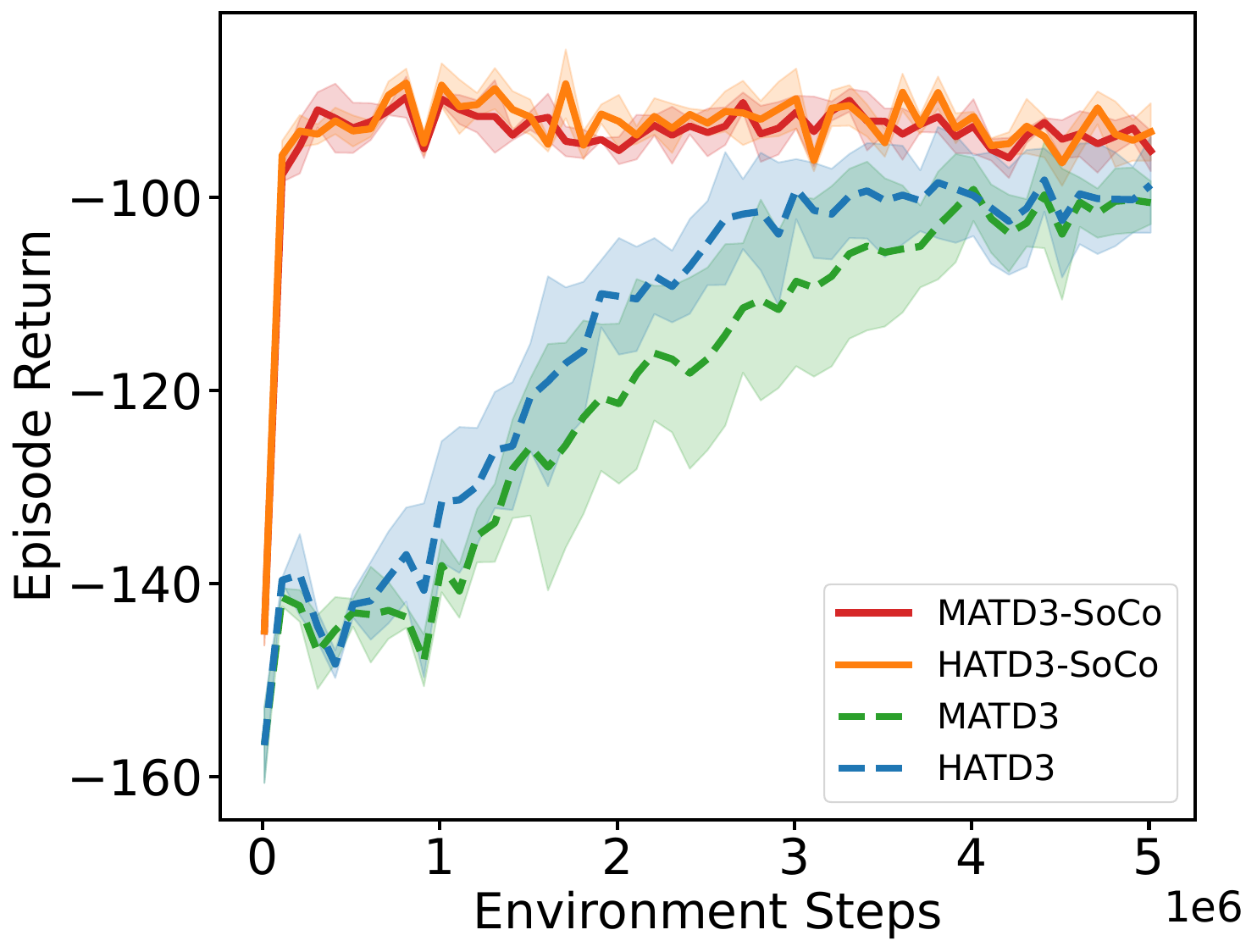}
        \caption{Spread (4 agents).}\label{fig:s4}
    \end{subfigure}
    \begin{subfigure}[b]{0.3\columnwidth}
        \centering
        \includegraphics[width=.9\textwidth]{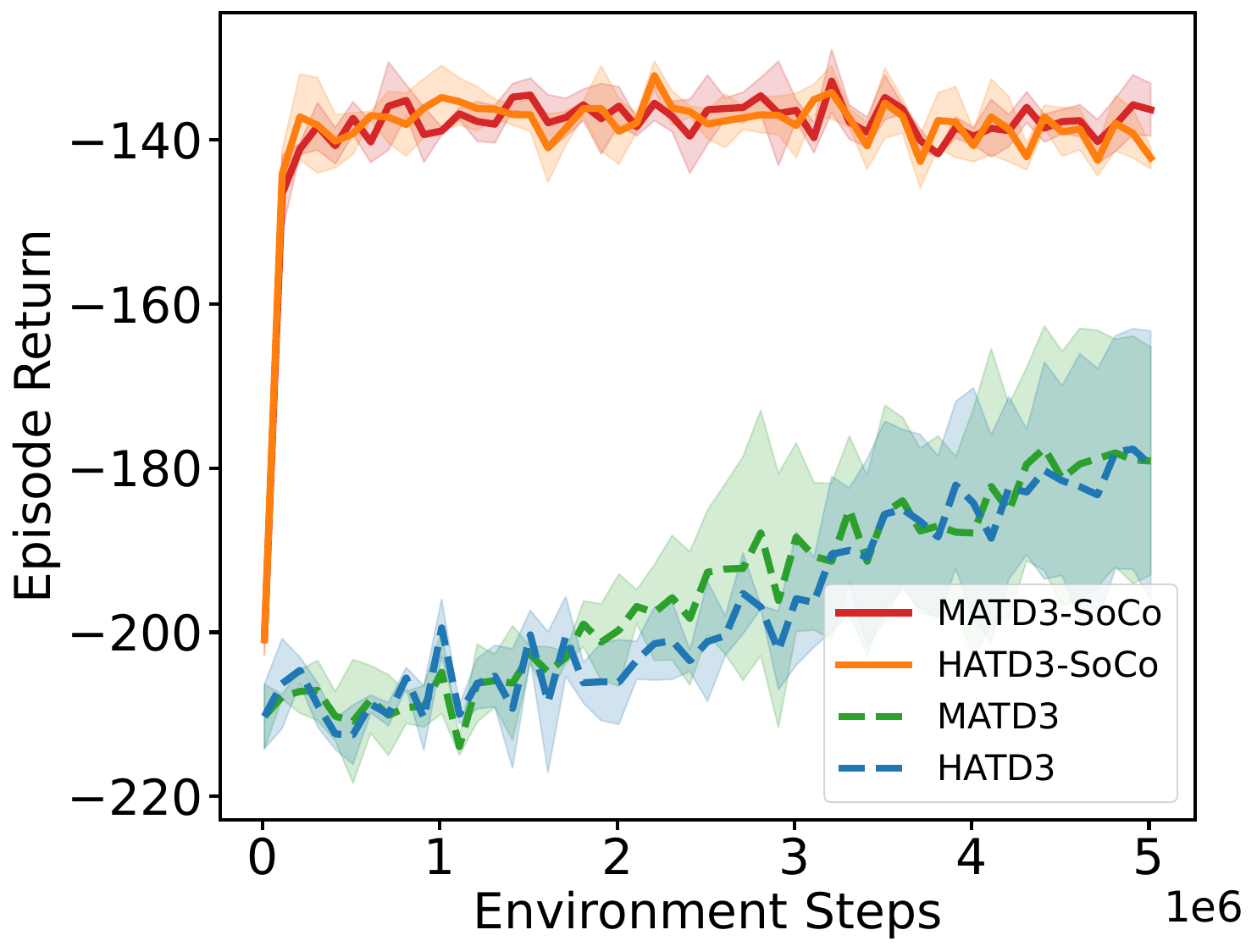}
        \caption{Spread (5 agents).}\label{fig:s5}
    \end{subfigure}\\
    \begin{subfigure}[b]{0.3\columnwidth}
        \centering
        \includegraphics[width=.9\textwidth]{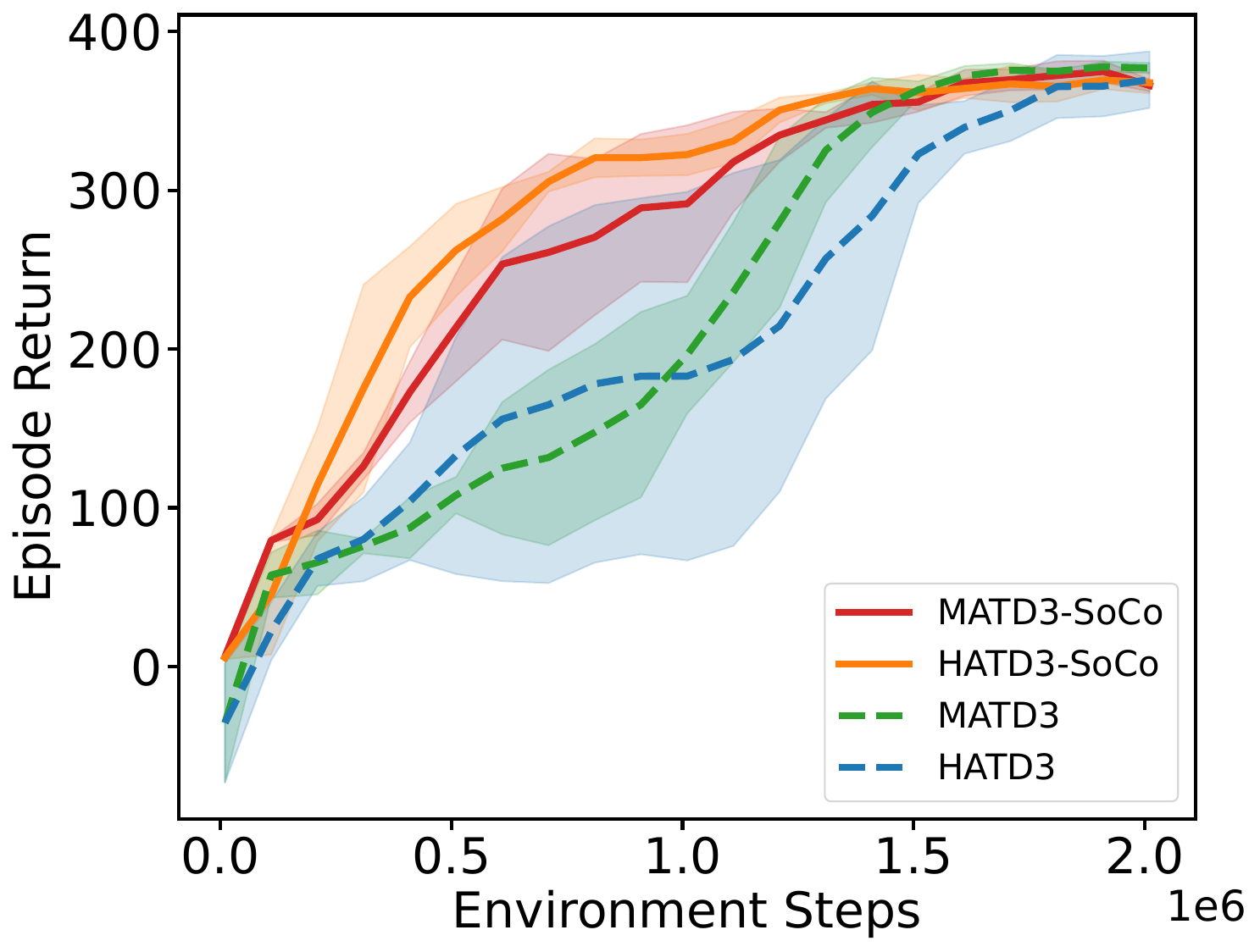}
        \caption{LongSwimmer (3 agents).}\label{fig:ls3}
    \end{subfigure}
    \begin{subfigure}[b]{0.3\columnwidth}
        \centering
        \includegraphics[width=.9\textwidth]{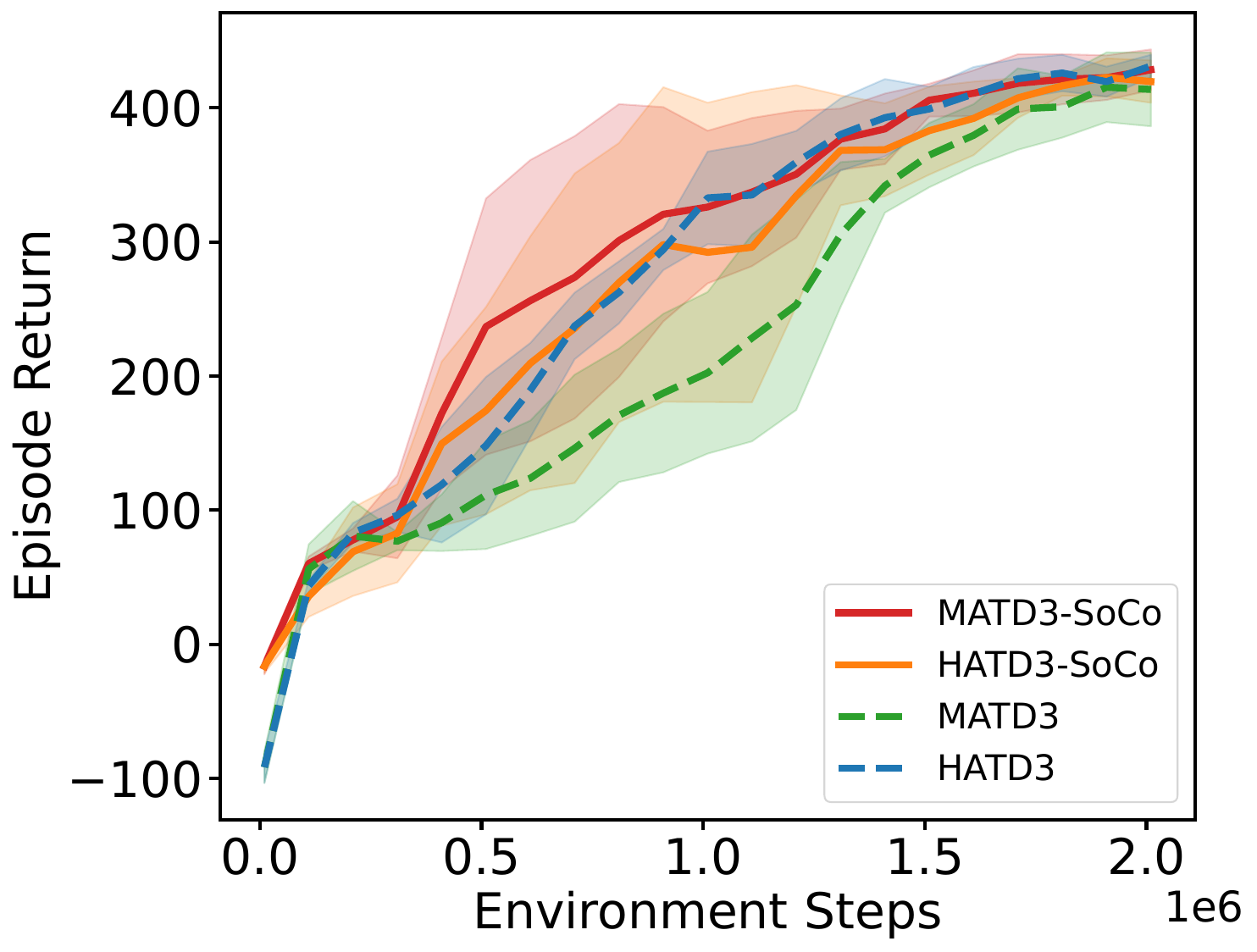}
        \caption{LongSwimmer (4 agents).}\label{fig:ls4}
    \end{subfigure}
    \begin{subfigure}[b]{0.3\columnwidth}
        \centering
        \includegraphics[width=.9\textwidth]{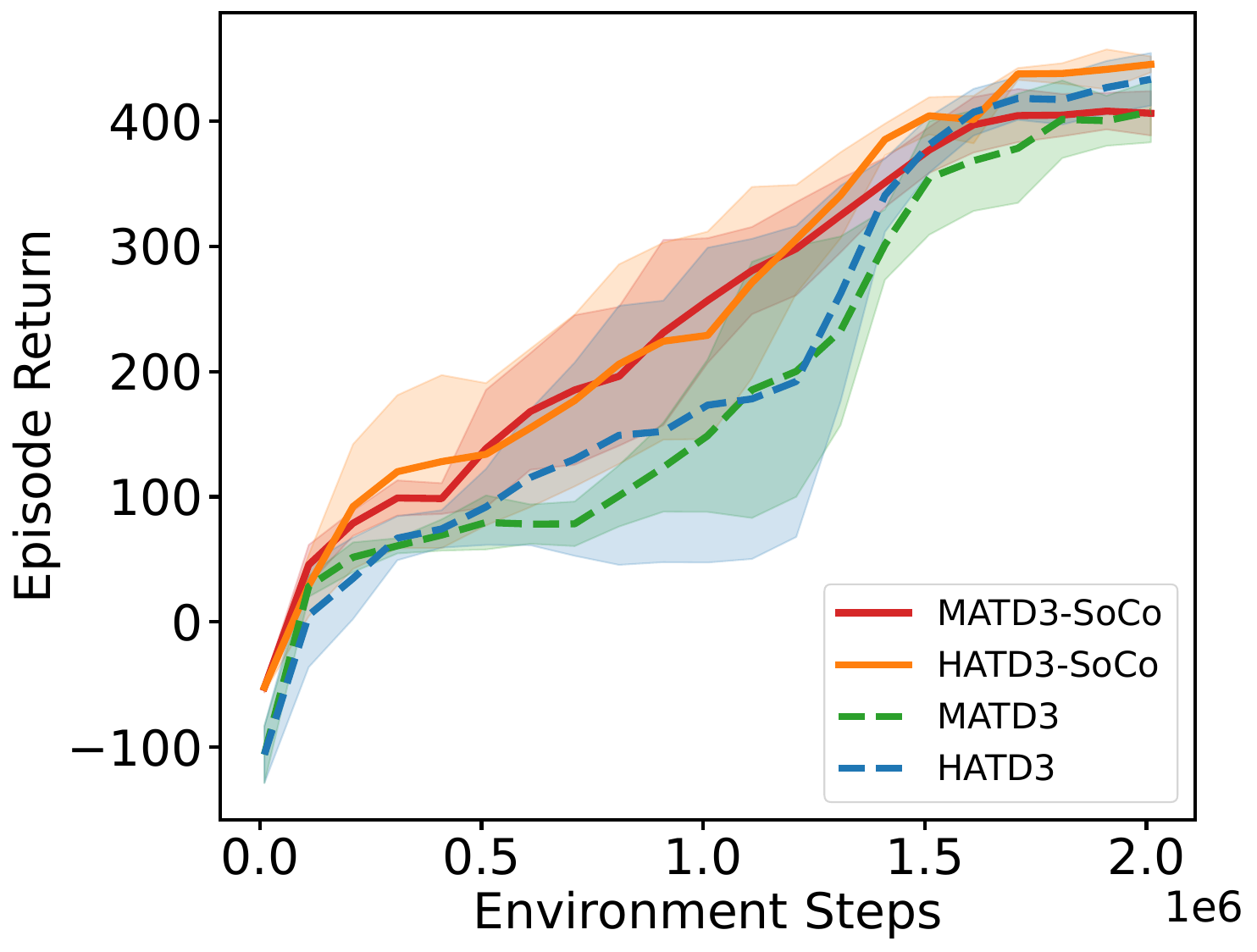}
        \caption{LongSwimmer (5 agents).}\label{fig:ls5}
    \end{subfigure}\\
    \begin{subfigure}[b]{0.3\columnwidth}
        \centering
        \includegraphics[width=.9\textwidth]{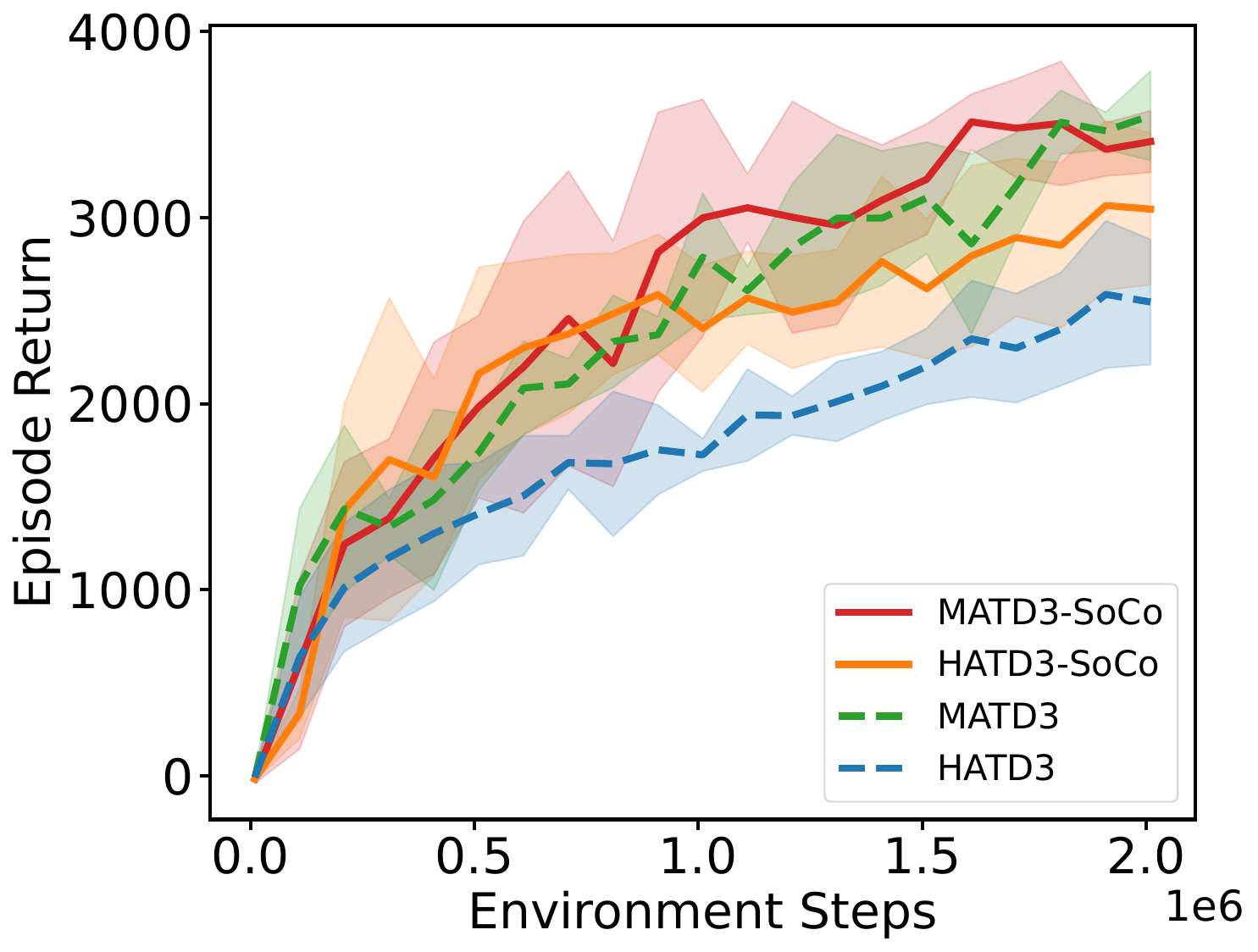}
        \caption{MultiHalfcheetah (2 agents).}\label{fig:h2}
    \end{subfigure}
    \begin{subfigure}[b]{0.3\columnwidth}
        \centering
        \includegraphics[width=.9\textwidth]{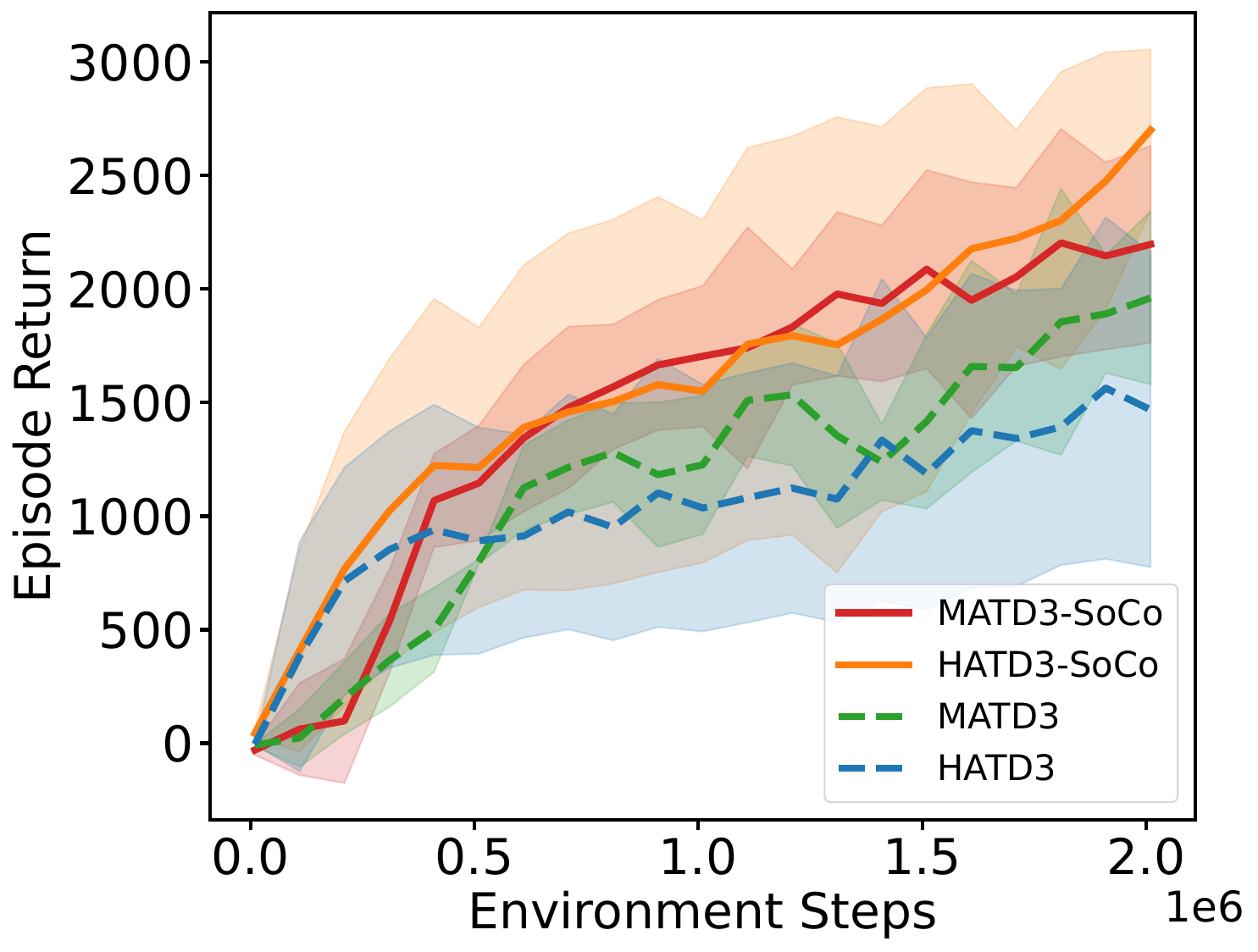}
        \caption{MultiHalfcheetah (3 agents).}\label{fig:h3}
    \end{subfigure}
    \begin{subfigure}[b]{0.3\columnwidth}
        \centering
        \includegraphics[width=.9\textwidth]{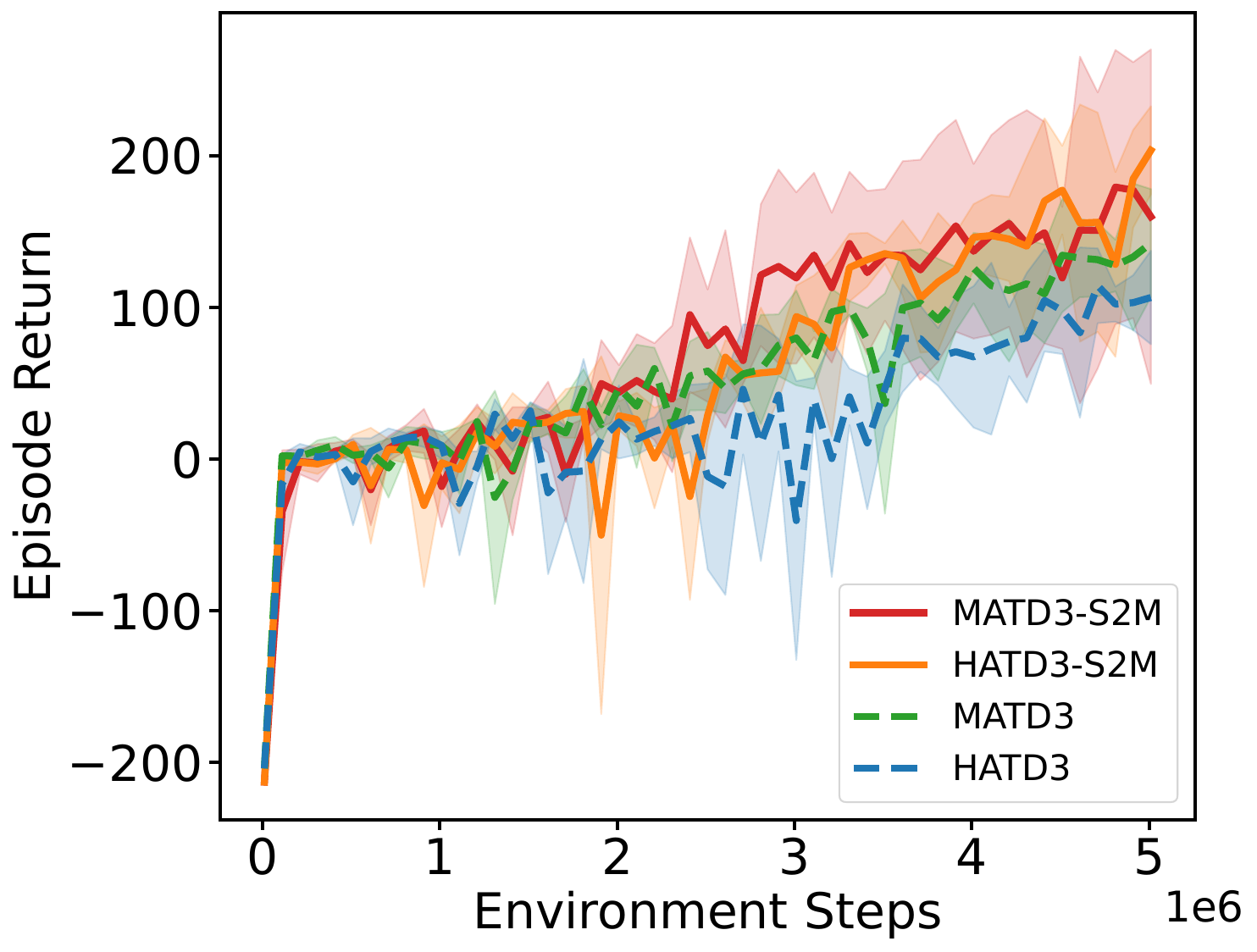}
        \caption{MultiWalker (2 agents).}\label{fig:m2}
    \end{subfigure}
    \caption{Training curves on nine tasks. Results are averaged over three random seeds, with solid and dashed lines indicating the mean performance and shaded areas representing one standard deviation.}
\label{fig:res}
\end{figure}
We evaluate SoCo on nine tasks across four scenarios with varying characteristics and difficulty. Across both backbone algorithms, SoCo improves training efficiency and achieves competitive or superior performance, demonstrating its effectiveness.

\paragraph{Spread.}
In the \textit{Spread} tasks, agents must learn not only to navigate to landmarks but also to resolve target assignment and avoid collisions, with difficulty growing rapidly as the number of agents increases. As shown in Figures \ref{fig:s3}–\ref{fig:s5}, training from scratch becomes highly inefficient under this setting. With SoCo, however, agents first acquire basic navigation skills from solo demonstrations, and during cooperative training, they only need to master target selection and collision avoidance via policy fusion. This significantly improves both training efficiency and final performance. For example, in the 5-agent task, SoCo converges faster and outperforms both backbone algorithms by more than $\textbf{20\%}$ in final performance, demonstrating its effectiveness in mitigating the challenge of goal ambiguity.

\paragraph{LongSwimmer.}In the \textit{LongSwimmer} task, agents collaboratively control a multi-segment worm to swim forward. As shown in Figures \ref{fig:ls3}-\ref{fig:ls5}, although the control difficulty is moderate and both backbone algorithms and SoCo eventually reach similar performance, our framework effectively speeds up training. For example, in the 3-agent task, HATD3-SoCo attains an average return of about $300$ at roughly $1.0$M steps, while vanilla HATD3 only reaches a comparable level around $1.6$M steps, saving nearly $\textbf{40\%}$ of training samples. These results highlight that SoCo successfully leverages solo demonstrations as a scalable prior to accelerate cooperative learning.

\paragraph{MultiHalfcheetah.}
The \textit{MultiHalfCheetah} task requires multiple HalfCheetah agents connected by elastic tendons to run forward in coordination. The tendon coupling already introduces instability, while the intrinsic difficulty of HalfCheetah control further compounds the challenge. Unlike \textit{Spread} or \textit{LongSwimmer}, this scenario also alters the agents’ mass, creating domain shifts that make solo policies non-transferable. Nevertheless, SoCo leverages action editor to adapt solo priors to the shifted dynamics while retaining their basic control skills. As shown in Figures \ref{fig:h2} and \ref{fig:h3}, this leads to markedly improved training efficiency for the backbone algorithms. In particular, on the 3-agent task, HATD3-SoCo improves the final performance by approximately $\textbf{83.91\%}$ over the backbone, highlighting the strength of SoCo in leveraging solo demonstrations to boost both efficiency and effectiveness of cooperative training.

\paragraph{MultiWalker.}
The MultiWalker task is the most challenging among the four scenarios. Agents must not only stabilize multiple walkers but also coordinate to carry a long, unstable package under noisy observations. The reward structure is harsh, and the domain shifts are severe, making direct transfer highly difficult. In this setting, backbone MARL algorithms struggle to learn effective package transport within the training budget. By contrast, SoCo leverages policy fusion to refine solo priors and adapt them to unstable multi-agent dynamics, enabling faster discovery of transport strategies and yielding clear gains in both training speed and final performance. Notably, as shown in \ref{fig:m2}, SoCo improves the final performance by $\textbf{91.51\%}$ on HATD3 and $\textbf{11.97\%}$ on MATD3 compared to their vanilla counterparts. This shows that SoCo can transfer solo knowledge even under extreme conditions, substantially improving both training efficiency and performance.

\subsection{Ablation Study}
\begin{figure}[!t]
\centering
    \begin{subfigure}[b]{0.3\columnwidth}
        \centering
        \includegraphics[width=.9\textwidth]{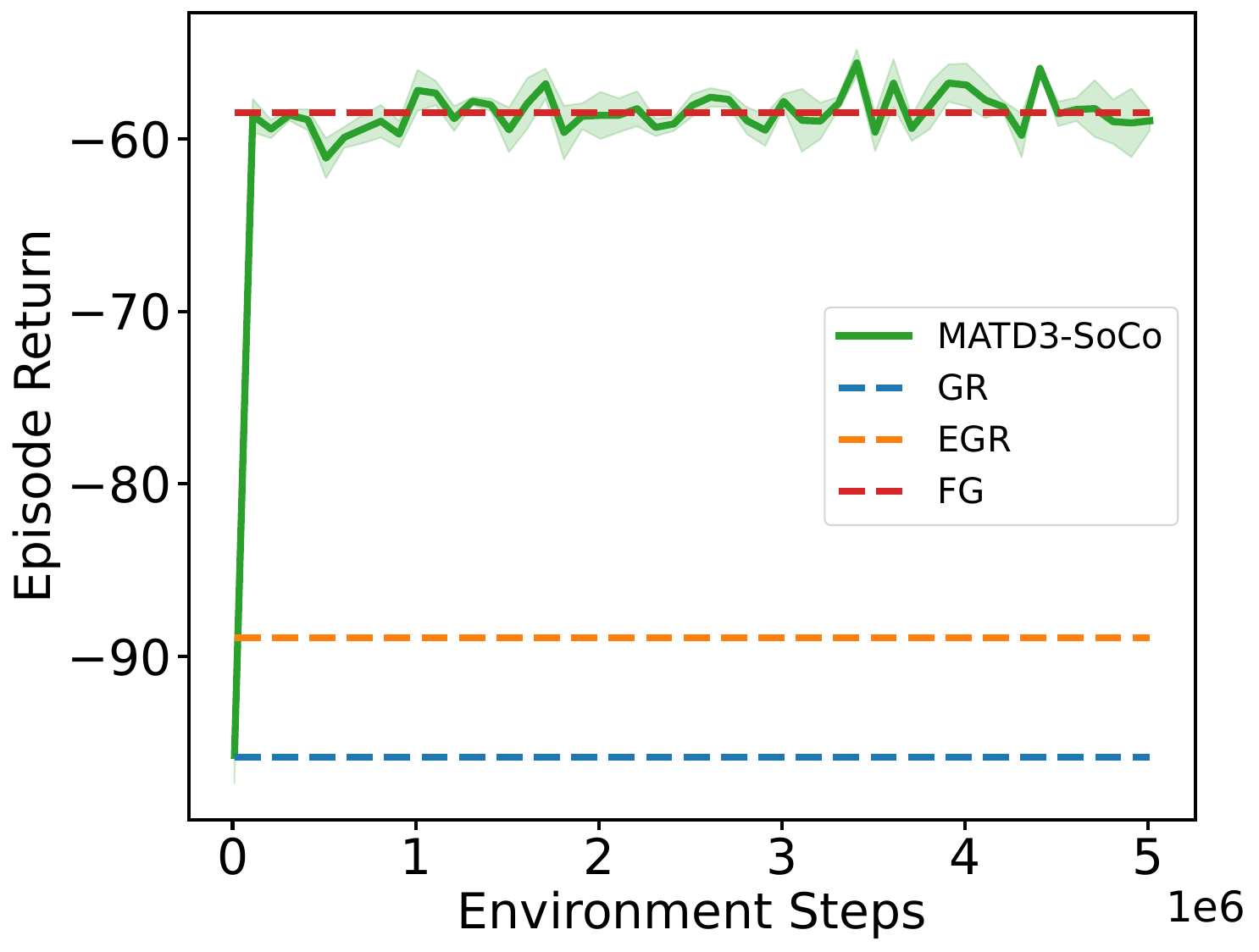}
        \caption{Ablation on Gating Selector.}\label{fig:ab_gate}
    \end{subfigure}
    \begin{subfigure}[b]{0.3\columnwidth}
        \centering
        \includegraphics[width=.9\textwidth]{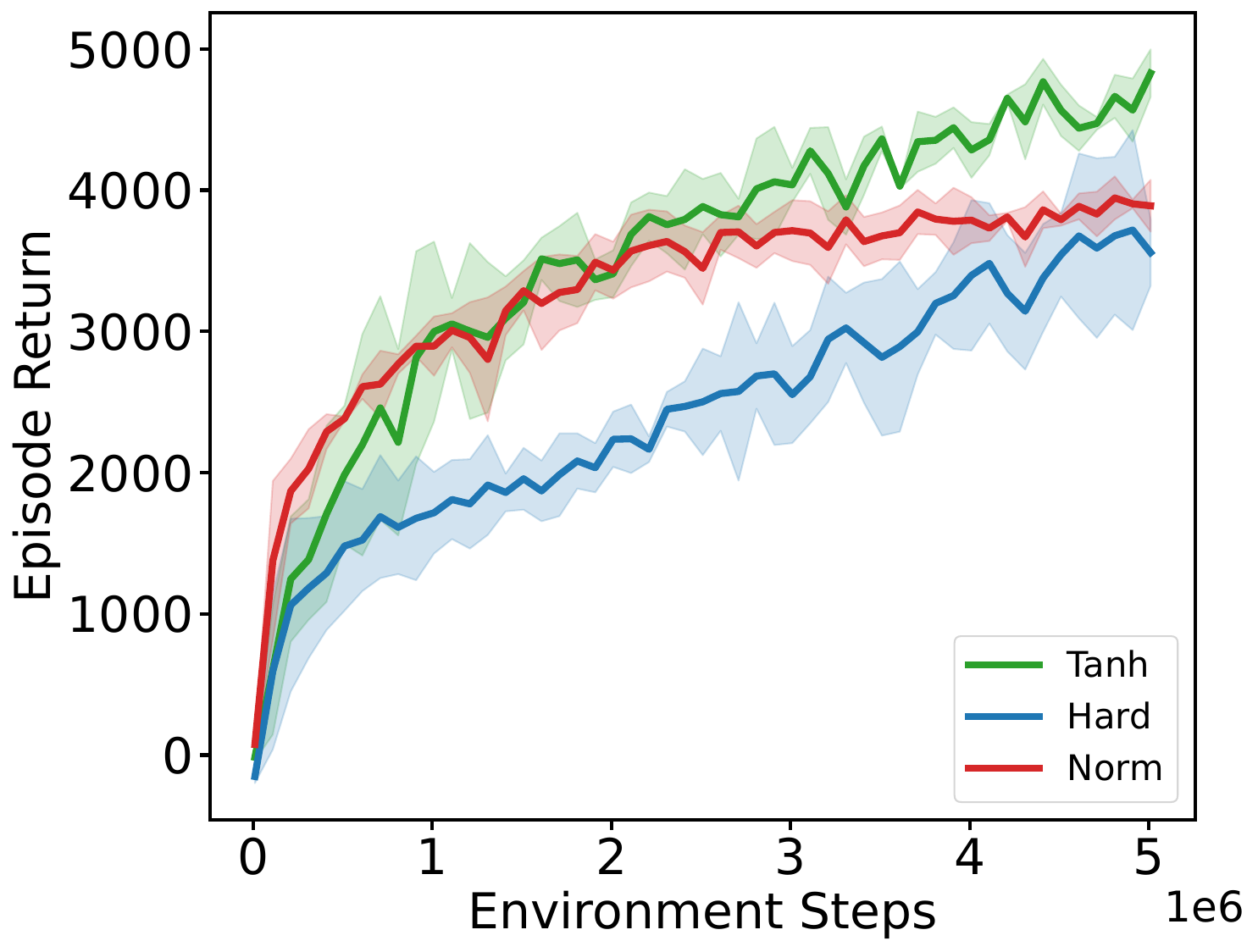}
        \caption{Ablation on Clip Operator.}\label{fig:ab_clip}
    \end{subfigure}
    \begin{subfigure}[b]{0.3\columnwidth}
        \centering
        \includegraphics[width=.9\textwidth]{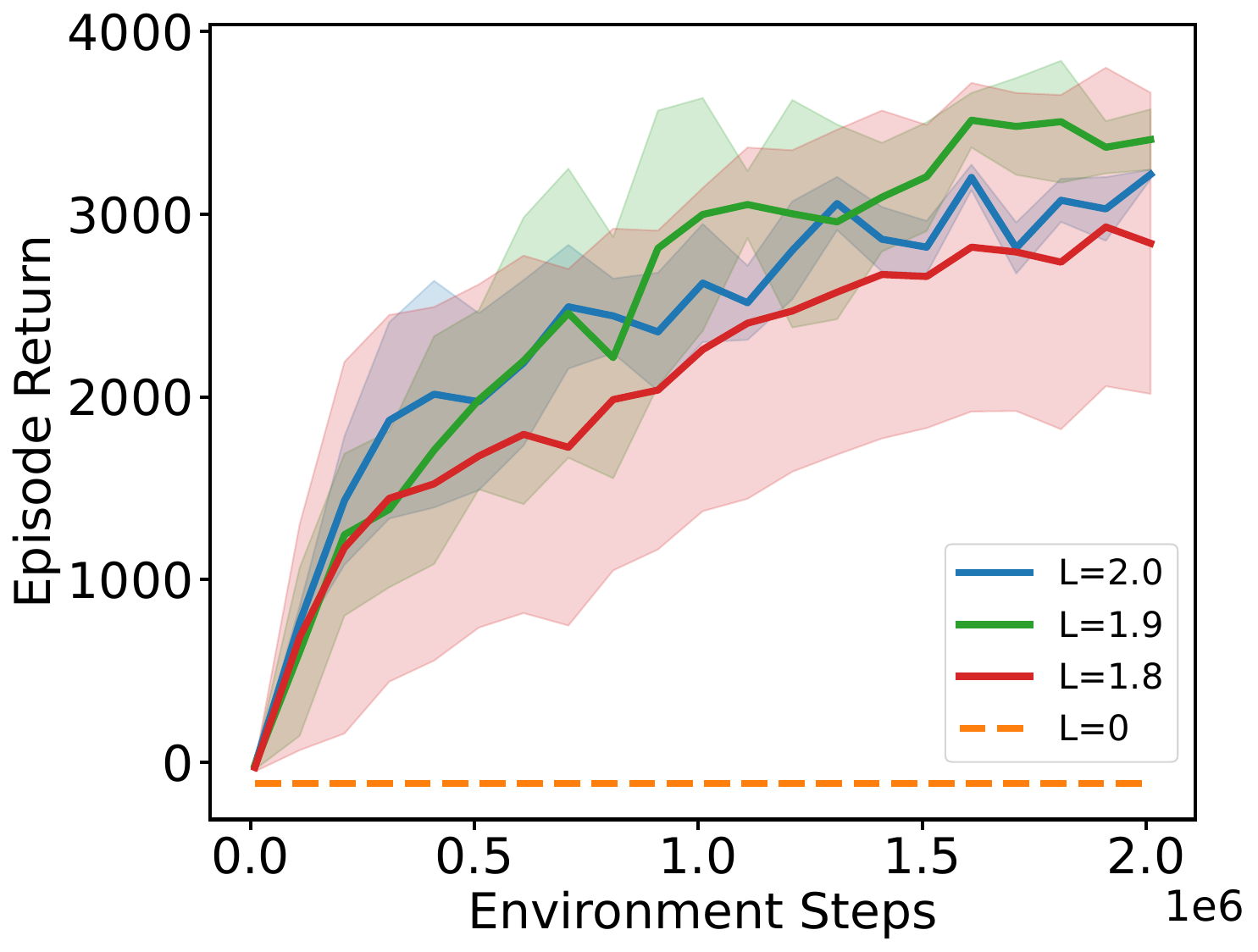}
        \caption{Ablation on Strength $L$.}\label{fig:ab_l}
    \end{subfigure}
    \caption{Ablation study of SoCo. Results are averaged over three random seeds, with solid and dashed lines indicating the mean performance and shaded areas representing one standard deviation.}
\label{fig:ab_res}
\end{figure}
\subsubsection{Component Ablation}

\paragraph{Gating Selector.}We conduct an ablation study on the 3-agent \textit{Spread} task to isolate the effect of the gating selector. Given the environment structure, setting the correction strength to zero ($L=0$) already yields strong performance, so we focus exclusively on the gating component. We compare three variants: (i) \textbf{Random Gating (RG)}, where targets are sampled randomly at each step; (ii) \textbf{Episode-wise Random Gating (ERG)}, where targets are randomly fixed at the start of each episode; and (iii) \textbf{Fixed Gating (FG)}, where \textit{distinct} targets are deterministically assigned by agent index at the beginning of each episode, serving as an oracle assignment in this scenario. As shown in Figure~\ref{fig:ab_gate}, randomized gating (RG / ERG) suffers from frequent target conflicts and poor coordination, whereas our learned gating selector can avoid conflicts and perform competitively to FG.

\paragraph{Clip Operator.}
As discussed in Section~\ref{sec:ae}, we adopt a $\tanh$-based clip operator to prevent fused actions from exceeding valid ranges. Nevertheless, SoCo is designed as a general framework, and different clipping strategies can be customized for specific tasks. To examine this flexibility and assess the suitability of our choice, we conduct experiments on the 2-agent \textit{MultiHalfCheetah} task, evaluating how alternative operators affect both training efficiency and final performance. Using MATD3 as the backbone, we compare two variants:

(i) \textbf{Norm}, which normalizes the action as $\mathrm{Clip}(\tilde{a}_t^i + \Delta a_t^i) = (\tilde{a}_t^i + \Delta a_t^i)/(L+1)$;

(ii) \textbf{Hard}, which directly truncates actions via $\mathrm{clamp}(\tilde{a}_t^i + \Delta a_t^i, -1, 1)$. 

As shown in Figure~\ref{fig:ab_clip}, the Norm operator accelerates early learning but suffers from weak asymptotic performance, as normalization continuously shrinks the effective action magnitude and reduces policy expressiveness. The Hard operator, on the other hand, truncates actions abruptly, suppressing gradient signals and leading to slow and unstable training. In contrast, our \textbf{$\tanh$}-based design achieves a smoother balance between boundedness and gradient flow, since gradients are only compressed near the action boundaries. This enables both stable learning dynamics and stronger final performance, making the $\tanh$-based operator a natural and effective default choice for SoCo. That said, the framework remains flexible to alternative operators when required by task dynamics.

\subsubsection{Hyperparameter Sensitivity}
An important hyperparameter in the SoCo framework is the correction strength $L$, which controls the degree to which the algorithm leverages knowledge from solo demonstrations. We conduct experiments on the 2-agent \textit{MultiHalfCheetah} task with $L \in \{0, 1.8, 1.9, 2.0\}$. Since this environment does not involve multi-goal settings, we can effectively isolate the influence of the gating selector and focus on the impact of this hyperparameter on SoCo’s performance. The results in Table~\ref{fig:ab_l} show that when $L$ is too small, SoCo relies excessively on solo demonstrations, which limits its training efficiency. In contrast, when $L$ is too large, SoCo adapts quickly to environmental changes in the early stage, but insufficient use of solo knowledge makes it difficult to discover better strategies later, leading to suppressed final performance. Thus, an appropriate choice of $L$ is essential for maximizing the effectiveness of SoCo. In addition, we also examined the case of $L=0$, where solo policies are directly applied in the multi-agent environment. The results reveal that domain shift prevents the agents from being successfully controlled, underscoring the necessity of policy fusion.

\section{Related Work}
\paragraph{MARL.} Multi-Agent Reinforcement Learning (MARL) has advanced rapidly, giving rise to diverse paradigms. Fully decentralized methods train and execute policies without centralized information \cite{tampuu2017multiagent,de2020independent}, but often suffer from limited coordination. By contrast, the Centralized Training with Decentralized Execution (CTDE) paradigm \cite{matd3,qmix,yu2021surprising,harl,ctde_new} has become dominant, enabling centralized training for coordination while preserving decentralized execution. In this paper, we focus on deterministic policy gradient methods under CTDE.

\paragraph{Transferable MARL.} To mitigate the high cost of training from scratch, transferable MARL aims to reuse experience from source tasks to accelerate learning in target tasks with limited interaction. Existing approaches include offline-to-online \cite{zhong2025offline}, multi-task \cite{chen2024variational,hissd,jha2025crossenvironment}, ad-hoc teamwork \cite{zhang2023fast,li2025llm}, and mixed-component or personalized data \cite{wang2023dm2,yu2025beyond}. While these methods broaden MARL’s applicability, they still assume sufficient multi-agent data. In contrast, exploiting solo demonstrations, abundant yet lacking cooperative signals, remains underexplored. Our work fills this gap by showing that such data can be effectively leveraged to improve both the training efficiency and performance of cooperative learning.

More detailed discussions are provided in Appendix~\ref{app:rw}.

\section{Conclusion}
In this paper, we studied an underexplored problem: how to exploit solo demonstrations to accelerate MARL. We propose a novel Solo-to-Collaborative RL (SoCo) framework, which leverages solo demonstrations by pretraining a shared solo policy and adapting it during cooperative training through policy fusion with a gating selector and an action editor. Experiments across diverse tasks show that SoCo improves training efficiency and achieves competitive or even superior performance, {highlighting that solo demonstrations provide a scalable and effective complement to multi-agent data, making cooperative learning more practical and broadly applicable.}

\bibliography{plain}
\bibliographystyle{plain}

\newpage
\appendix
\renewcommand{\appendixpagename}{\LARGE Technical Appendices}
\appendixpage
\startcontents[section]
\printcontents[section]{l}{1}{\setcounter{tocdepth}{2}}
\newpage
\section{Detailed Related Work} \label{app:rw}
\paragraph{MARL.} Multi-Agent Reinforcement Learning (MARL) has advanced rapidly in recent years, giving rise to diverse paradigms and methods. Fully decentralized approaches train and execute policies without centralized information \cite{tampuu2017multiagent,de2020independent}, but their performance is often constrained by the absence of communication among agents. By contrast, the Centralized Training with Decentralized Execution (CTDE) paradigm \cite{oliehoek2008optimal, matignon2012coordinated, IntroCTDE, ctde_new} has become the mainstream, enabling agents to learn with centralized information for coordination while still executing policies in a decentralized manner. Representative algorithms include HASAC \cite{liu2024maximum}, HARL \cite{harl}, MAPPO \cite{yu2021surprising}, QMIX \cite{qmix}, and MATD3 \cite{matd3}. In this work, we adopt deterministic policy gradient methods within the CTDE paradigm, with particular focus on MATD3 and HATD3.
\paragraph{Transferable MARL.} Since training MARL from scratch is often sample-inefficient and costly, transferable MARL seeks to reuse experience from source settings to accelerate learning in target tasks with limited additional interaction. Existing approaches span several directions: offline-to-online MARL \cite{zhong2025offline}, which leverages offline pretraining to speed up online exploration and correct distributional shift; multi-task MARL \cite{hu2021updet, wang2020rode, zhang2023odis, chen2024variational, hissd, jha2025crossenvironment}, which extracts transferable knowledge from multiple source tasks and applies it to unseen ones; ad-hoc teamwork \cite{stone2010ad,zhang2023fast,li2025llm}, which exposes agents to diverse teammates to improve robustness when coordinating with unseen partners; and MARL with mixed-component or personalized data \cite{wang2023dm2,yu2025beyond}, which constructs datasets from individual trajectories generated by different cooperative policies, enriching training diversity while preserving per-step consistency. While these methods broaden the applicability of MARL, they all rely on sufficient multi-agent data. In contrast, the potential of exploiting solo demonstrations, abundant but lacking cooperative signals, remains largely unexplored. Our work addresses this gap by showing that such data can be effectively leveraged to accelerate cooperative training, thereby opening a promising new avenue.
\newpage
\section{Algorithm Pseudocode}\label{app:algo}
\begin{algorithm}[!htpb]
   \caption{Solo-to-Collaborative Reinforcement Learning (SoCo)}  
   \label{alg:s2m}
\begin{algorithmic}
   \STATE {\bfseries Input:} Datasets of solo demonstration $\mathcal{D}$ and edit strength $L$.
   \STATE Initialize the parameters $w$ for solo policy $\beta_w$, $\phi=\{\varphi,\theta\}$ for weight assigner $g_\varphi$ and coordination policy $\pi_\theta$, $\{\psi_j\}_{j=1}^2$ for $\{Q_j\}_{j=1}^2$, and $\bar\psi_1,\bar\psi_2, \bar\phi$ for target networks.
   \STATE Train solo policy $\beta_w$ with $\mathcal{D}$ according to Eq. (\ref{eq:bc_loss}). 
   \STATE Initialize the replay buffer $\mathcal{B}$.
   \FOR{$i=1${ \bfseries to} $T_{\text{max}}$}
   \STATE Obtain the joint observation $\textbf{o}_t$ from the environment.
   \STATE \textcolor{blue}{// Agent-wise Solo-to-Collaborative Transfer}
   \FOR{$n=1${ \bfseries to} $N$}
   \STATE \textcolor{blue}{// Observation Decomposition}
   \STATE Decompose local observation $o_t^n$ into solo views $\{o_t^{n,k}\}_{k=1}^{G_n}$.
   \STATE \textcolor{blue}{// Policy Fusion}
   \STATE Calculate $\boldsymbol{a}_t^n=\beta_w(\{\boldsymbol{o}_t^n\})$ and obtain solo action $\tilde{a}_t^n$ by Eq. (\ref{eq:gate})
   \STATE Calculate editing action $\Delta{a}_t^n$ by Eq. (\ref{eq:aedit}).
   \STATE Obtain final action $a_t^n$ by combining $\tilde{a}_t^n$ and $\Delta{a}_t^n$ according to Eq. (\ref{eq:comb}).
   \ENDFOR
   \STATE \textcolor{blue}{// Cooperative MARL Training}
   \STATE Use $\boldsymbol{a}_t=(a_t^1,\dots,a_t^N)$ to interact with the environment and save $(s_t,\boldsymbol{o}_t,\boldsymbol{a}_t,r_t,\boldsymbol{o}_{t+1})$ into $\mathcal{B}$.
   
   \STATE Sample a batch of transitions $\{(s_t,\boldsymbol{o}_t,\boldsymbol{a}_t,r_t,\boldsymbol{o}_{t+1})\}$ from $\mathcal{B}$. 
   \STATE Update critics $Q_1,~Q_2$ and fused policy $\Pi_\phi$ through standard MARL algorithms.
   \ENDFOR 
\end{algorithmic}
\end{algorithm}
\newpage
\section{Experiments Details}\label{app:exp_details}
\subsection{Environments}\label{app:env}

\begin{figure}[!ht]
\centering
    \begin{subfigure}[b]{0.3\columnwidth}
        \centering
        \includegraphics[width=.7\textwidth]{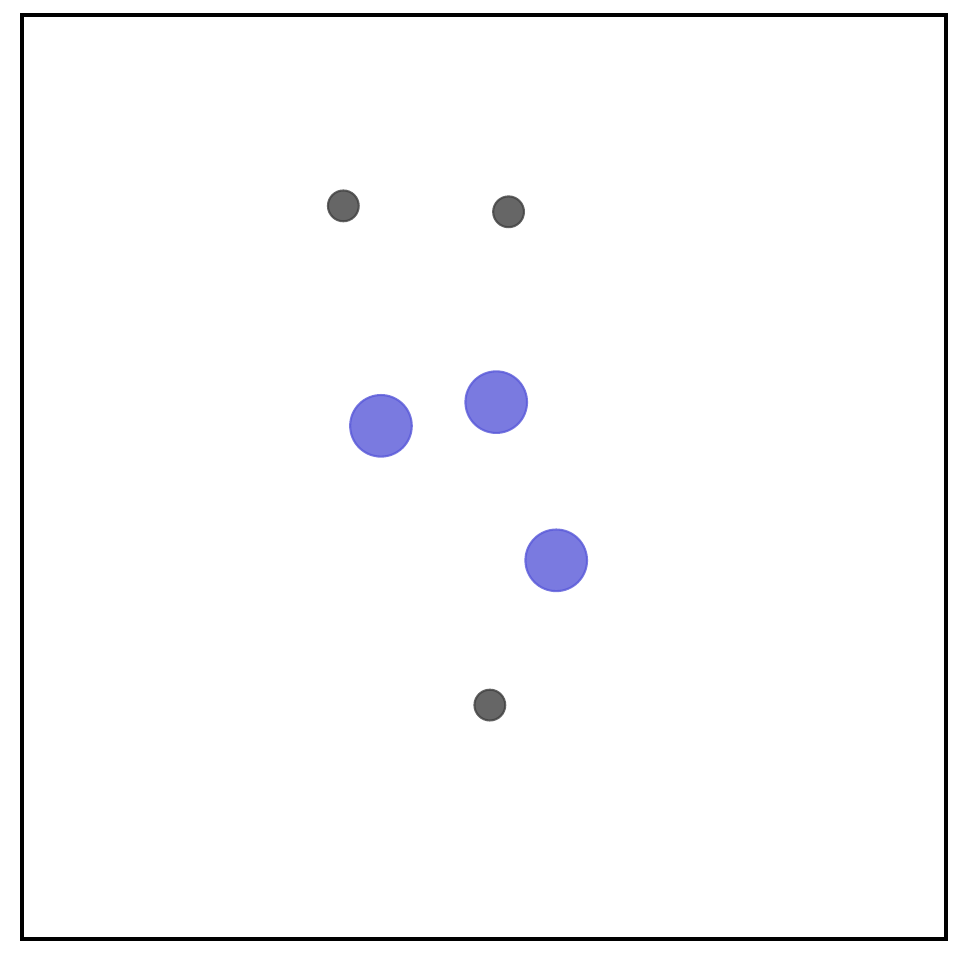}
        \caption{Spread (3 agents).}\label{fig:es3}
    \end{subfigure}
    \begin{subfigure}[b]{0.3\columnwidth}
        \centering
        \includegraphics[width=.7\textwidth]{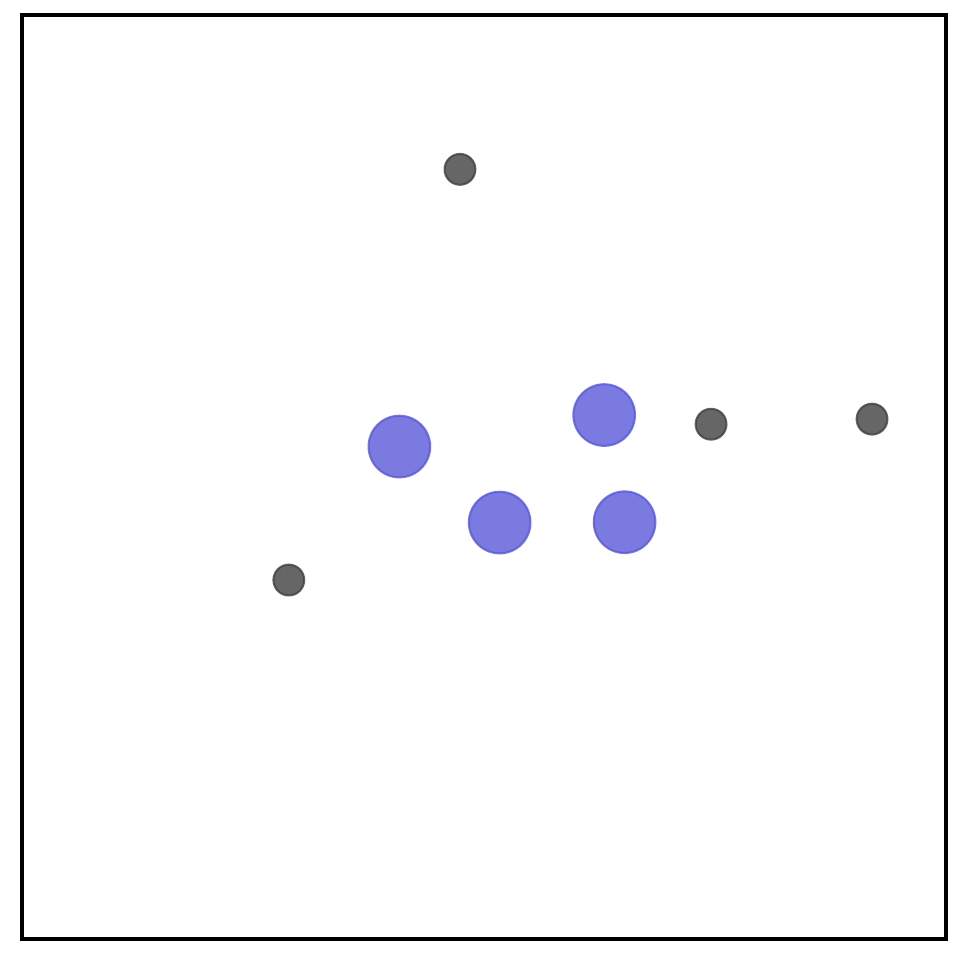}
        \caption{Spread (4 agents).}\label{fig:es4}
    \end{subfigure}
    \begin{subfigure}[b]{0.3\columnwidth}
        \centering
        \includegraphics[width=.7\textwidth]{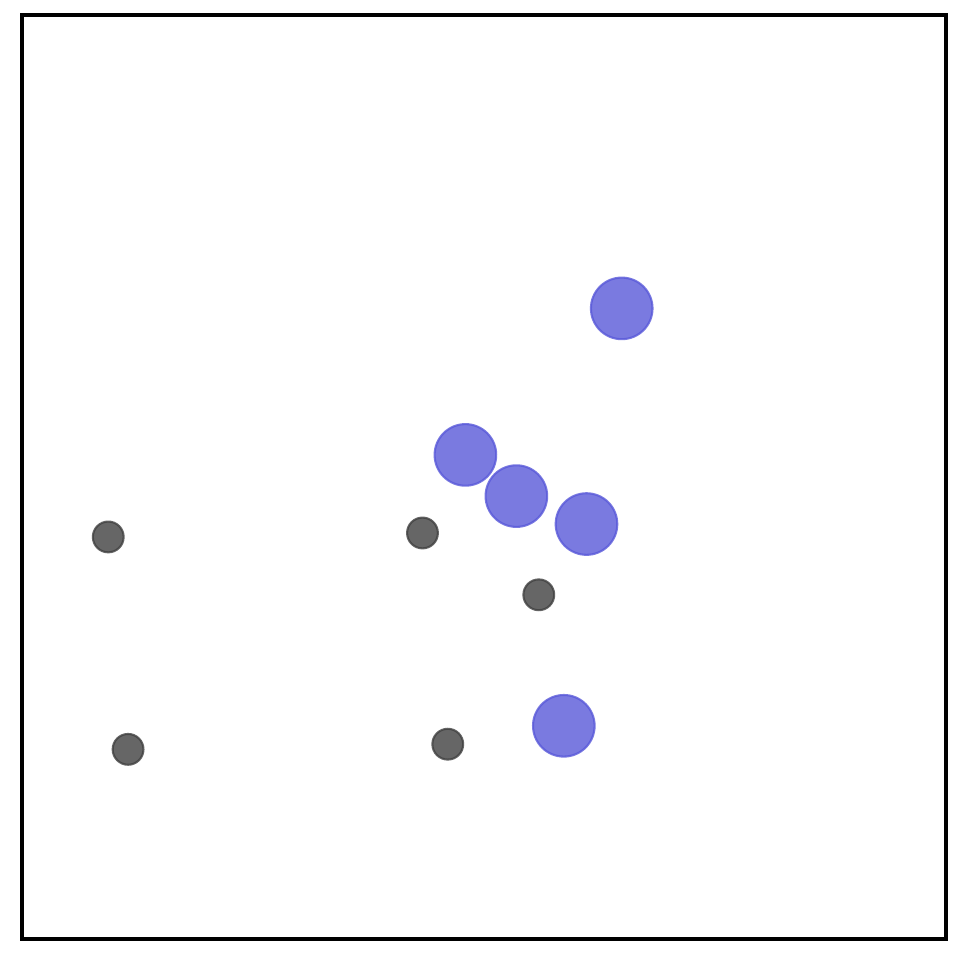}
        \caption{Spread (5 agents).}\label{fig:es5}
    \end{subfigure}\\
    \begin{subfigure}[b]{0.3\columnwidth}
        \centering
        \includegraphics[width=.8\textwidth]{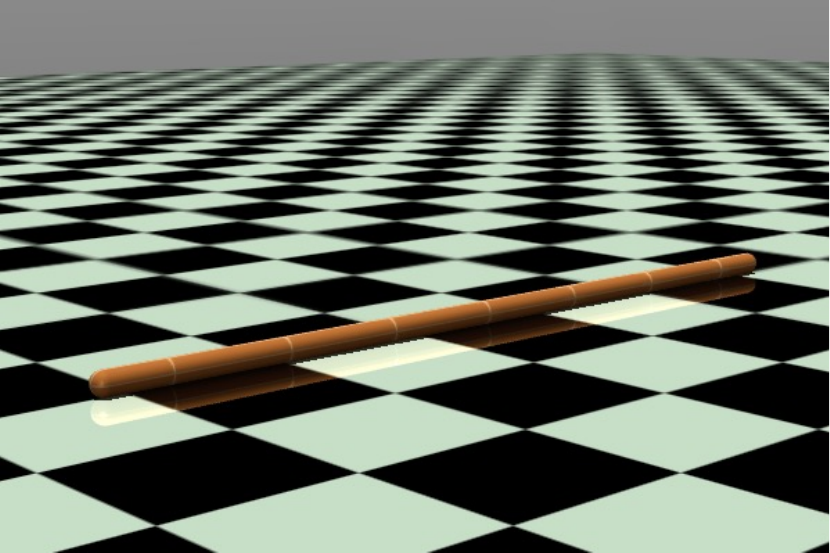}
        \caption{LongSwimmer (3 agents).}\label{fig:els3}
    \end{subfigure}
    \begin{subfigure}[b]{0.3\columnwidth}
        \centering
        \includegraphics[width=.8\textwidth]{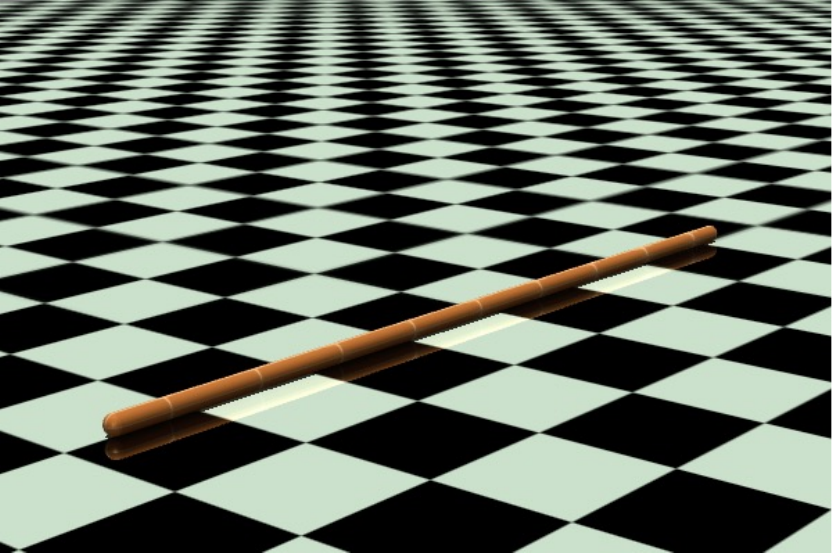}
        \caption{LongSwimmer (4 agents).}\label{fig:els4}
    \end{subfigure}
    \begin{subfigure}[b]{0.3\columnwidth}
        \centering
        \includegraphics[width=.8\textwidth]{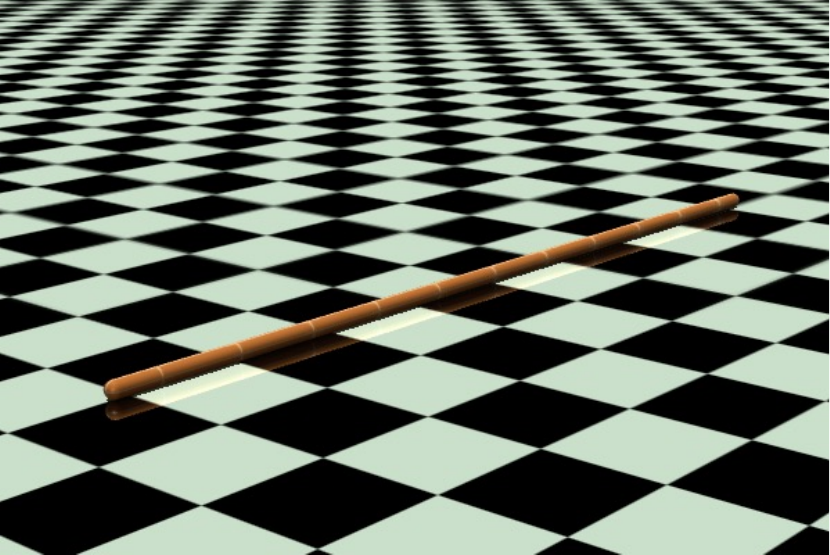}
        \caption{LongSwimmer (5 agents).}\label{fig:els5}
    \end{subfigure}\\
    \begin{subfigure}[b]{0.3\columnwidth}
        \centering
        \includegraphics[width=.8\textwidth]{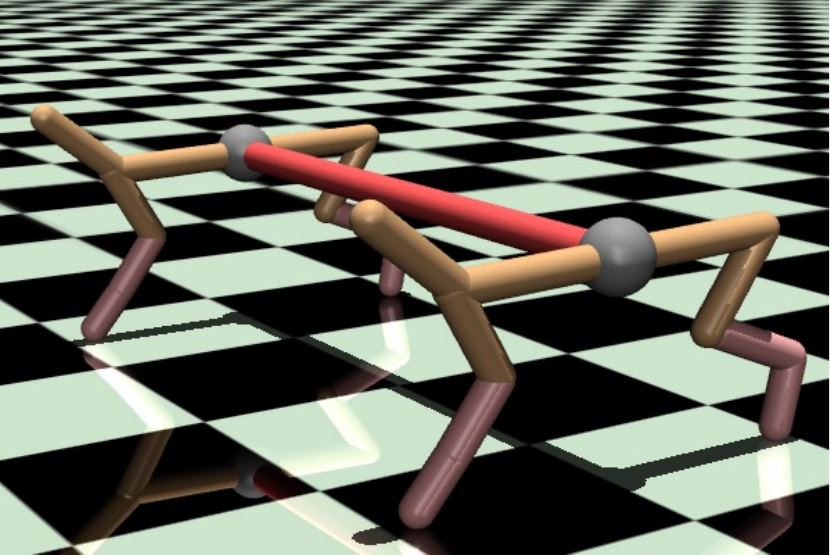}
        \caption{MultiHalfcheetah (2 agents).}\label{fig:eh2}
    \end{subfigure}
    \begin{subfigure}[b]{0.3\columnwidth}
        \centering
        \includegraphics[width=.8\textwidth]{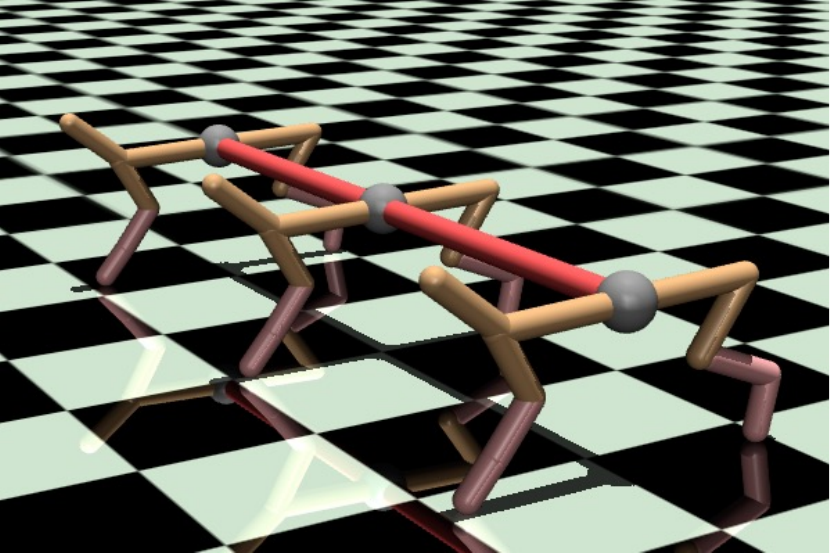}
        \caption{MultiHalfcheetah (3 agents).}\label{fig:eh3}
    \end{subfigure}
    \begin{subfigure}[b]{0.3\columnwidth}
        \centering
        \includegraphics[width=.8\textwidth]{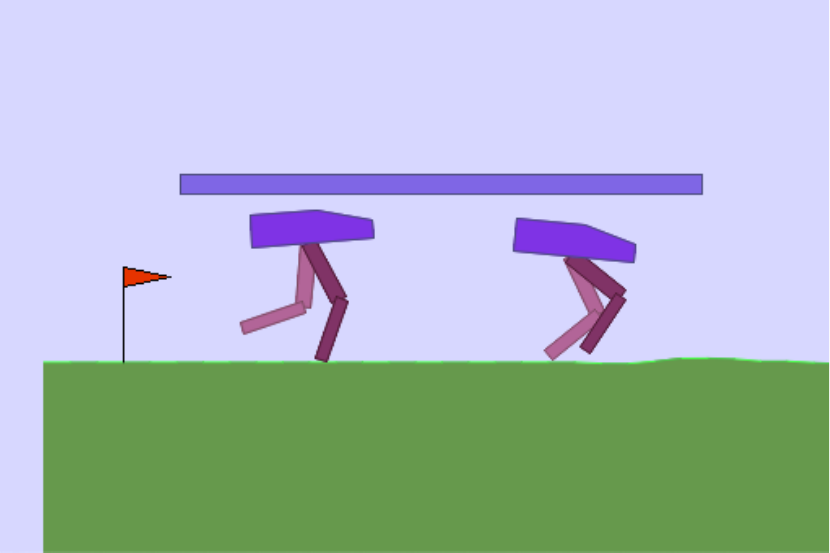}
        \caption{MultiWalker (2 agents).}\label{fig:em2}
    \end{subfigure}
    \caption{All the cooperative tasks in our experiments.}
\label{fig:env}
\end{figure}

We evaluate SoCo on nine tasks across four representative cooperative scenarios:
\paragraph{Spread \protect\cite{lowe2017multi,terry2021pettingzoo}.} As shown in Figures \ref{fig:es3}-\ref{fig:es5}, in this environment, $N$ agents are initialized at random positions in a bounded 2D plane, while $K=N$ landmarks are also randomly placed without overlap. Agents must navigate to distinct landmarks while avoiding collisions. The per-step reward for each agent $i$ is defined as the average of a global and a local component:

$$
r_t^i = \tfrac{1}{2}\big(r_t^{\text{global}} + r_t^{\text{local}, i}\big).
$$

The global reward is shared across agents and encourages coverage of landmarks:

$$
r_t^{\text{global}} = -\sum_{k=1}^K \min_{j \in \mathcal{N}} \| p_t^j - l_k \|_2,
$$

where $p_t^j$ is the position of agent $j$, and $l_k$ is the position of landmark $k$.

The local reward penalizes collisions:

$$
r_t^{\text{local}, i} =
\begin{cases}
   -C_t^i, & \text{if agent $i$ collides with $C_t^i$ other agents}, \\
   0, & \text{otherwise}.
\end{cases}
$$

Finally, the environment reward is the sum over all agents’ individual rewards:

$$
R_t = \sum_{i \in \mathcal{N}} r_t^i.
$$

We evaluate on tasks with 3, 4, and 5 agents.

\paragraph{LongSwimmer \protect\cite{facmac,gymnasium_robotics2023github}.} 
As shown in Figures~\ref{fig:els3} and \ref{fig:els5}, in this environment, a $(2N+1)$-segment worm must be controlled to swim forward. Each pair of adjacent segments is connected by a joint, and each agent is responsible for controlling two consecutive joints in sequence. The worm’s initial state is sampled from a uniform distribution within a predefined range, while its initial velocity is drawn from Gaussian noise to diversify the dynamics. The per-step reward for each agent $i$ is:

$$
r_t^i = v_t - 0.0001\cdot\sum_{i\in\mathcal{N}}\|a_t^i\|_2^2,
$$

where $v_t$ is the forward velocity of the worm, $a_t^i$ is the action taken by agent $i$.

The environment reward is defined as the average of all agents’ rewards:

$$
R_t = \frac{1}{N}\sum_{i \in \mathcal{N}} r_t^i.
$$

We evaluate on tasks with 3, 4, and 5 agents.

\paragraph{MultiHalfCheetah \protect\cite{facmac,gymnasium_robotics2023github}.} 
As shown in Figures \ref{fig:eh2} and \ref{fig:eh3}, in this environment, $N$ HalfCheetah agents are connected in series by elastic tendons and must collaboratively run forward. Each agent’s initial state is sampled from a uniform distribution within a predefined range, and its initial velocity is drawn from Gaussian noise to diversify dynamics. The per-step reward for each agent $i$ is:

$$
r_t^i = v_t^i - 0.1\cdot\|a_t^i\|_2^2,
$$

where $v_t^i$ is the forward velocity of agent $i$, $a_t^i$ is its action.

The environment reward is defined as the average of all agents’ rewards:

$$
R_t = \frac{1}{N}\sum_{i \in \mathcal{N}} r_t^i.
$$

We evaluate on tasks with 2 and 3 HalfCheetahs.

\paragraph{MultiWalker \protect\cite{gupta2017cooperative, terry2021pettingzoo}.}

As shown in Figure \ref{fig:em2}, in this environment, $N$ bipedal robots must collaboratively lift and carry a long package forward. The terrain has a randomly undulating profile at the start of each episode. Walkers are initialized at fixed, equally spaced positions in standing poses; to diversify initial conditions, a small random external force is applied to each walker’s head at $t=0$. The package length scales proportionally with the number of walkers, and each walker’s observation is corrupted with noise.

At each step, each walker receives a progress reward equal to the forward displacement of the package, plus a small shaping penalty for head tilting and a $-10$ penalty if a walker falls:

$$
r_t^{i} \;=\; \Delta x_t^{\text{package}} \;-\; 5 \cdot \Delta \theta_t^{\text{head}, i} \;-\; 10\cdot \mathbf{1}\{\text{walker } i \text{ falls}\}.
$$

Episodes terminate if the package falls, leaves the left edge, or if any walker falls, in which case all walkers receive $-100$. If the package exits the right edge, termination occurs with reward $0$. 

The environment reward at each step is the sum of individual rewards:

$$
R_t \;=\; \sum_{i \in \mathcal{N}} r_t^i.
$$

We evaluate on task with 2 walkers.

\subsection{Solo Demonstration}\label{app:solo}

\begin{figure}[!ht]
\centering
    \begin{subfigure}[b]{0.24\columnwidth}
        \centering
        \includegraphics[width=.6\textwidth]{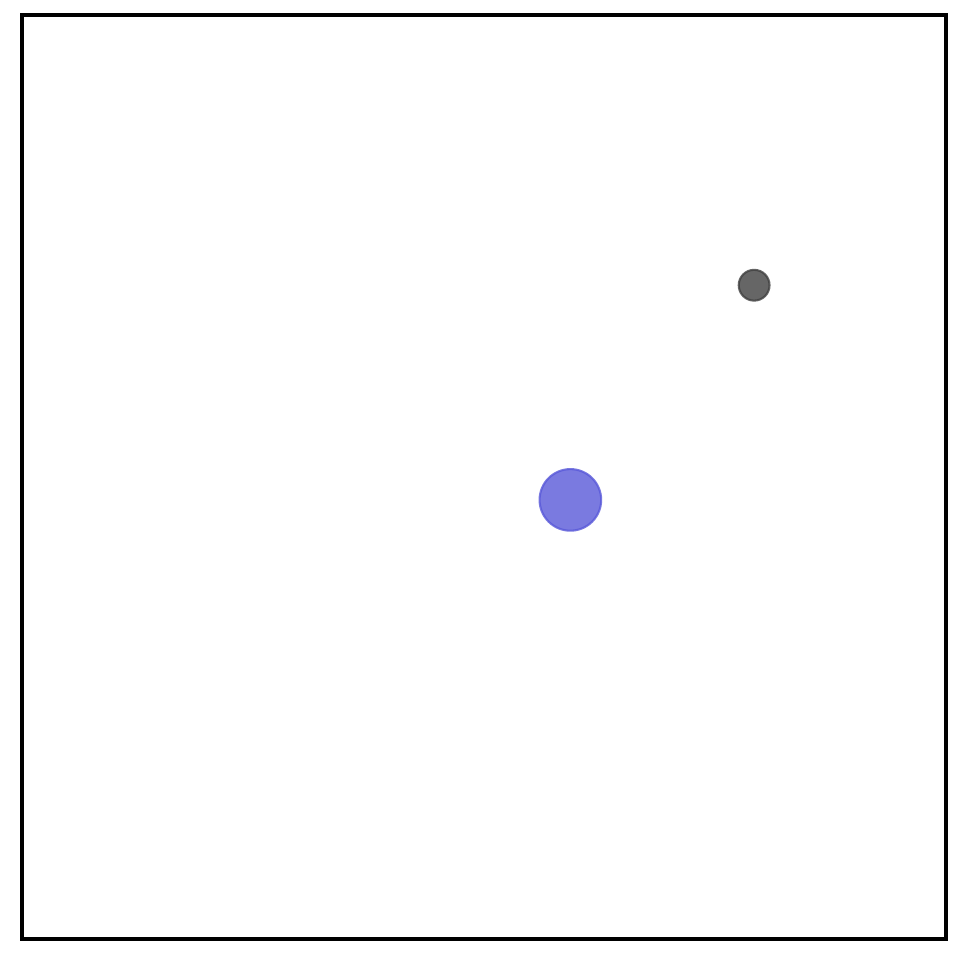}
        \caption{Spread.}\label{fig:es1}
    \end{subfigure}
    \begin{subfigure}[b]{0.24\columnwidth}
        \centering
        \includegraphics[width=.9\textwidth]{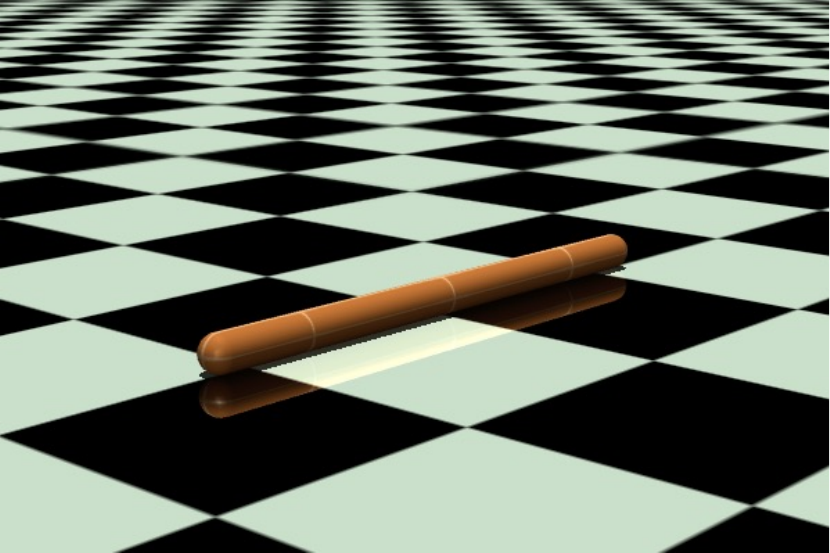}
        \caption{LongSwimmer.}\label{fig:els1}
    \end{subfigure}
    \begin{subfigure}[b]{0.24\columnwidth}
        \centering
        \includegraphics[width=.9\textwidth]{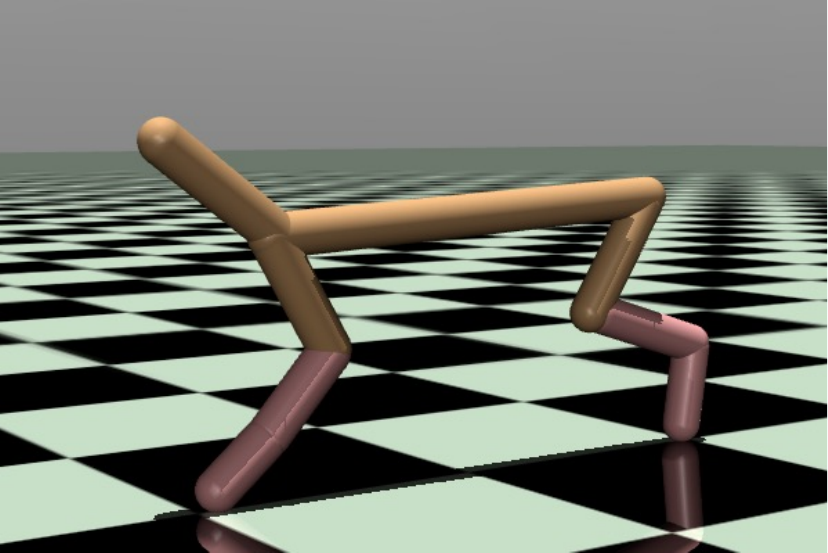}
        \caption{MultiHalfcheetah.}\label{fig:eh1}
    \end{subfigure}
    \begin{subfigure}[b]{0.24\columnwidth}
        \centering
        \includegraphics[width=.9\textwidth]{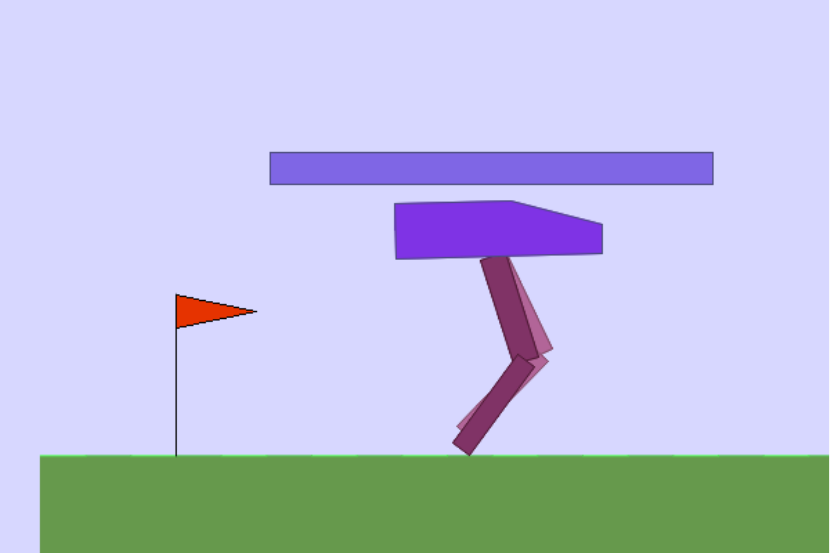}
        \caption{MultiWalker.}\label{fig:em1}
    \end{subfigure}
    \caption{Solo tasks corresponding to each cooperative scenario.}\label{fig:solo_env}
\end{figure}

\subsubsection{Data Collection}
For each cooperative scenario, we first train a policy on its corresponding solo task using TD3 \cite{td3}, and then collect $1$M transitions to learn the solo policy. Table \ref{tab:demo} summarizes the average episode returns of the solo task demonstrations.

\begin{table}[ht]
\centering
\caption{Average episode return of collected solo demonstrations.}
\label{tab:demo}
\begin{tabular}{lcccc}
\toprule
\textbf{Scenario} & \textbf{Spread} & \textbf{LongSwimmer} & \textbf{MultiHalfcheetah} & \textbf{MultiWalker} \\
\midrule
\textbf{Average Episode Return} &-14.77 &119.44  & 7054.87 & 197.18\\
\bottomrule
\end{tabular}
\end{table}
\subsubsection{Solo Tasks vs. Cooperative Scenarios}

These solo tasks, illustrated in Figure \ref{fig:solo_env}, exhibit noticeable gaps from their cooperative counterparts, ranging from goal ambiguity (\textit{Spread}), to moderate domain shift (\textit{LongSwimmer}), to notable domain shift and cooperative difficulty (\textit{MultiHalfCheetah}), and to severe domain shift with substantial cooperative difficulty (\textit{MultiWalker}).

Specifically, in the \textit{Spread} scenario, the solo task allows an agent to observe only a single target, whereas in the cooperative setting, multiple targets are visible simultaneously. In the \textit{LongSwimmer}, the motion of the worm is affected by the actions of other agents, introducing a moderate domain shift. In \textit{MultiHalfCheetah}, the solo task doubles the agent’s mass and removes tendon constraints, making it simpler than the coupled cooperative case. Finally, in \textit{MultiWalker}, the solo task differs drastically from the cooperative environment: the package length and walker positions change, observations are noisy, and interference from teammates is absent in solo but present in multi-agent training, resulting in severe domain shift and substantially higher cooperative difficulty.

\subsection{Implementation Details}\label{app:imp}
Our implementation and experiments are based on the HARL codebase \cite{harl}. The additional components introduced by SoCo, i.e., the solo policy, gating selector, and action editor, share the same architecture as the backbone actor network, implemented as 2-layer MLPs with ReLU activations. For action fusion, we adopt a $\tanh$-based clip operator: when $\Delta a \equiv 0$, no constraint is applied; otherwise, the fused action is bounded through a $\tanh$ transformation. We use Adam \cite{kingma2014adam} for optimization. Additionally, in the 3-agent \textit{MultiHalfCheetah} environment, the tendon structure can destabilize the MuJoCo simulator. To mitigate this, we impose an additional constraint on the output of HATD3-SoCo, clipping it to the range $[-0.85, 0.85]$.
\subsection{Hyperparameters}\label{app:hyper}
Except for the correction strength $L$ in SoCo, all hyperparameters follow the default or recommended (when available) settings in HARL to ensure fair comparison. The detailed configurations are reported in Table \ref{tab:hyper}. 
\begin{table}[ht]
\centering
\caption{Shared hyperparameters for all algorithms.}
\label{tab:hyper}
\begin{tabular}{lclc}
\toprule
\textbf{Hyperparameter} & \textbf{Value} & \textbf{Hyperparameter} & \textbf{Value} \\
\midrule
Batch Size & $1000$ & Buffer Size & $1000000$ \\
Hidden Size & $256$ ($128$ for \textit{Spread}) & Discount Factor $\gamma$ & $0.99$\\
$n$-step TD & $10$ ($1$ for \textit{Spread}) & Explore Noise & $0.1$\\
Policy Noise & $0.2$ & Noise Clip & $0.5$ \\
Policy Delay & $2$ & Soft Update Coefficient & 0.005 \\
 Actor Learning Rate & 0.0005 & Critic Learning Rate &0.001\\
\bottomrule
\end{tabular}
\end{table}

For SoCo, $L$ is an important hyperparameter that controls the extent to which knowledge from solo demonstrations is leveraged. The values of $L$ used for each task and backbone algorithm are summarized in Table \ref{tab:l_values}. Different tasks require different $L$ values, as the optimal balance depends on factors such as the degree of domain shift and the inherent difficulty of the cooperative environment.

\begin{table}[ht]
\centering
\caption{Correction strength $L$ used in SoCo for each task and backbone algorithm.}
\label{tab:l_values}
\begin{tabular}{lcc}
\toprule
\textbf{Task} & \textbf{MATD3-SoCo} & \textbf{HATD3-SoCo} \\
\midrule
Spread-3    & 0 & 0 \\
Spread-4    & 0 & 0 \\
Spread-5    & 0 & 0 \\
LongSwimmer-3    & 3.15 & 2.20 \\
LongSwimmer-4    & 3.10 & 2.90\\
LongSwimmer-5    & 2.10 & 2.85 \\
MultiHalfCheetah-2 & 1.90 & 2.00 \\
MultiHalfCheetah-3 & 1.90 & 1.90 \\
MultiWalker-2 & 2.00 & 2.00 \\
\bottomrule
\end{tabular}
\end{table}

\end{document}